\documentclass{article}
\usepackage[numbers, compress]{natbib} % natbib must be defined here, not in macros
\usepackage[final]{neurips_2024}
\usepackage[dvipsnames]{xcolor} % xcolor must be defined here, not in macros
\usepackage{microtype}
\usepackage{graphicx}  
\usepackage{subfigure} 
\usepackage{booktabs}
\usepackage{hyperref}
\usepackage{algorithm}
\usepackage{algcompatible}
\usepackage{amsmath}
\usepackage{soul}
\usepackage{xspace}
\usepackage{dsfont}
\usepackage{bm}
\usepackage{bbm}
\usepackage{nicefrac}       % compact symbols for 1/2, etc.
\usepackage[scaled=.8]{beramono} % reduce \texttt size
\usepackage{amsthm}
\usepackage{amsfonts}
\usepackage{amssymb}
\usepackage{array}
\usepackage{mathtools}
\usepackage{caption}
\usepackage{subcaption}
\usepackage{fontawesome}
\usepackage{thmtools}
\usepackage{thm-restate}
\usepackage{lipsum}
\usepackage{pifont}
\usepackage{makecell}
\usepackage{tablefootnote}
\usepackage[tight-spacing=true]{siunitx}
\usepackage{multirow}
\usepackage{afterpage}
\usepackage{tabularx} 
\usepackage{url}
\usepackage{tikz}
\usepackage{indentfirst}
\usepackage{enumitem}
\usepackage{epsdice}
\usepackage{wrapfig}
\usepackage[export]{adjustbox}
\usepackage{xparse}

\hypersetup{
  colorlinks=true,
  citecolor=Blue,
  linkcolor=Blue,
  urlcolor=Blue
}

\usepackage{cleveref}
\newcommand\sdots{\hbox to 1em{.\hss.\hss.}} % dots with smaller horizontal space
\DeclareMathAlphabet\mathbfcal{OMS}{cmsy}{b}{n} % Bold mathcal
\newcommand{\smallsum}{\textstyle\sum\nolimits}
\DeclareMathSymbol{\shortminus}{\mathbin}{AMSa}{"39}

\setitemize[0]{leftmargin=*}

\definecolor{mygreen}{RGB}{43, 138, 62}
\definecolor{myred}{RGB}{201, 42, 42}

\newcommand*\colourcheck[1]{%
  \expandafter\newcommand\csname #1check\endcsname{\textcolor{#1}{\ding{51}}}%
}
\colourcheck{mygreen}

\newcommand*\colourmark[1]{%
  \expandafter\newcommand\csname #1mark\endcsname{\textcolor{#1}{\ding{55}}}%
}
\colourmark{myred}

\newcommand{\Ours}{AutoGuide}
\title{AutoGuide: Automated Generation and \\ Selection of Context-Aware Guidelines for \\ Large Language Model Agents}

\author{
  Yao Fu $^{1}$\thanks{Equal contribution. } 
  \hspace{1cm}
  Dong-Ki Kim $^{2}$\footnotemark[1] 
  \hspace{1cm}
  Jaekyeom Kim $^{2}$
  \hspace{1cm}
  Sungryull Sohn $^{2}$
  \\
  \textbf{Lajanugen Logeswaran $^{2}$}
  \hspace{1cm}
  \textbf{Kyunghoon Bae $^{2}$}
  \hspace{1cm}
  \textbf{Honglak Lee $^{1,2}$} \\
  $^{1}$University of Michigan\qquad 
    $^{2}$LG AI Research
}
\begin{document}
\maketitle
\begin{abstract}
Recent advances in large language models (LLMs) have empowered AI agents capable of performing various sequential decision-making tasks.
However, effectively guiding LLMs to perform well in unfamiliar domains like web navigation, where they lack sufficient knowledge, has proven to be difficult with the demonstration-based in-context learning paradigm.
In this paper, we introduce a novel framework, called \textsc{\Ours}, which addresses this limitation by automatically generating context-aware guidelines from offline experiences.
Importantly, each context-aware guideline is expressed in concise natural language and follows a conditional structure, clearly describing the context where it is applicable.
As a result, our guidelines facilitate the provision of relevant knowledge for the agent's current decision-making process, overcoming the limitations of the conventional demonstration-based learning paradigm.
Our evaluation demonstrates that \textsc{\Ours} significantly outperforms competitive baselines in complex benchmark domains, including real-world web navigation.
\end{abstract}

\section{Introduction}\label{sec:intro}

Recent advances in large language models (LLMs) have empowered AI agents to address various sequential decision-making tasks and applications \citep{wang2023survey,xi2023rise}.
The foundation of these successes involves the planning and reasoning capabilities of pre-trained LLMs, enabling agents to execute effective policies \cite{brohan2023can,wei2022chain}.
The predominant approach to leveraging these (typically closed source) models for sequential decision making tasks is to provide demonstrations in the form of in-context examples.
However, direct application of this learning paradigm can be limited, especially in target domains where the LLM has insufficient prior knowledge such as in web navigation, where LLM agents generally achieve low success rates due to diverse and dynamic contents \cite{koh2024visualwebarena,deng2023mind2web,gur2023html,zhou2023webarena}.
Providing all available experiences as demonstrations to an agent can further be unsuccessful due to context length limitations, prompt sensitivity, and difficulty with complex reasoning~\citep{lu-etal-2022-fantastically,dong2022survey,min-etal-2022-rethinking,kaddour2023challenges}.

On the other hand, LLMs excel in interpreting concise instructions provided as natural language, an ability that is also reinforced in the instruction-tuning phase of LLMs.
Inpired by this, we explore data-driven strategies that leverage offline experiences to extract actionable knowledge to help guide LLM agents.
As offline experiences implicitly convey valuable knowledge about desirable and undesirable policies in domains, they promise to serve as a useful resource for improving an LLM agent's decision-making in situations where the pre-trained LLM lacks understanding.
Despite this potential benefit, a critical challenge lies in effectively extracting the implicit information embedded in offline data.
%%%%%%%%%

\begin{figure*}[t]
    \centering
    \includegraphics[width=\linewidth]{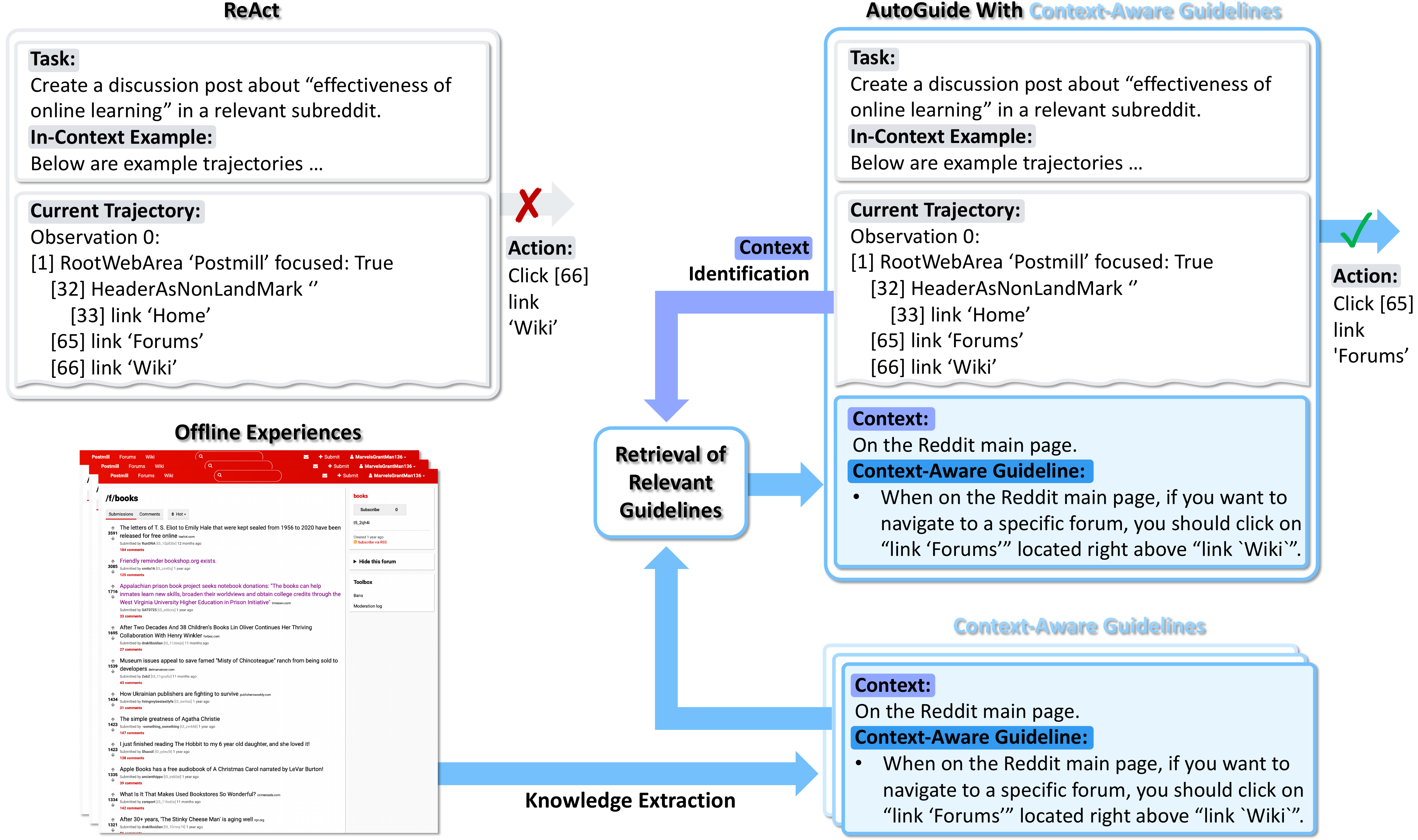}
    \vspace{-0.4cm}
    \caption{\textsc{\Ours} aims to extract the implicit knowledge embedded in offline experiences and help the decision-making process of an LLM agent. Specifically, our method generates a comprehensive set of context-aware guidelines from offline data and explicitly identifies when each guideline is applicable by generating its corresponding context. Our context-aware guidelines enable providing pertinent guidelines at test time by identifying the context of the current trajectory, leading to correct decision-making compared to baselines without context-aware guidelines.
    %\textcolor{red}{since we changed the webarena prompt, will also need to change the context and guideline, to this please: When on the main page of a Reddit-like platform, if you want to navigate to a specific forum, you should click on the "link 'Forums'", which is located at the early part of the observation, right above the link 'Wiki'. The correct action format to do this is ```click [link_id]```. }
    } 
    \label{fig:guideline-intro}
    \vspace{-0.6cm}
\end{figure*}

To address the challenge of extracting knowledge from offline data, we propose a novel framework, called \textsc{\Ours}. Specifically, \textsc{\Ours} automatically derives a comprehensive set of context-aware guidelines from offline experiences. Our method applies these context-conditional guidelines to enhance the performance of an LLM agent by retrieving guidelines relevant to the agent's current state and incorporating them into the prompt during testing (see \Cref{fig:guideline-intro}). Notably, we generate context-aware guidelines in concise natural language statements, effectively compressing knowledge in offline data. Moreover, context-aware guidelines clearly describe the contexts where they are applicable, so \textsc{\Ours} enables an LLM agent to select pertinent guidelines for its current decision-making process. As a result, \textsc{\Ours} achieves the highest success rates compared to competitive baselines in complex sequential decision-making benchmark environments.

%%%%%%%%%%%%%%%%%%%%%%%%%%%%%%%%%%%%%%%%%%%%%%%%%%%%%%%%%
%%%%%%%%%%%%%%%%%%%%%%%%%%%%%%%%%%%%%%%%%%%%%%%%%%%%%%%%%

\textbf{Our contribution.} In summary, we present the following main contributions in this paper:
\begin{itemize}[leftmargin=*, wide, labelindent=0pt, topsep=0pt]
    \itemsep 0pt 
    \item \textbf{Principled method based on context-aware guidelines (\Cref{sec:method}):} We develop two modules to automatically generate context-aware guidelines from offline experiences: the context identification module for identifying the context of a given trajectory, and the guideline extraction module for extracting a desired guideline corresponding to that context. The outcome is a set of domain knowledge in concise natural language that enhances decision-making by providing pertinent information.
    \item \textbf{Comprehensive evaluation of \textsc{\Ours} (\Cref{sec:main_result}):} We show \textsc{\Ours}'s capability in extracting helpful context-aware guidelines in various interactive benchmark domains, including navigating real-world web domains (e.g., GitHub). Our results highlight the effectiveness of \textsc{\Ours}, which significantly outperforms baselines without context-aware guidelines.
    \item \textbf{Analyses with important perspectives (\Cref{sec:analysis}):} We study various aspects of \textsc{\Ours}, such as the significance of determining the applicability of each guideline based on generated contexts. 
    We also investigate the generalization ability of context-aware guidelines and demonstrate that our guidelines enhance the performance across out-of-domain tasks.
\end{itemize}

\section{Related Work}
% \textcolor{red}{todo for lajan: if you could make necessary changes according to the feedback we received for ICML, that would be great :)} -> WIP

\textbf{LLM-based agents.}
Language models have recently been shown to possess strong priors for sequential decision-making tasks, which has given rise to LLM-powered agents \citep{wang2023survey,xi2023rise,zheng2023seeact,zeng2023agenttuning}.
Agents need to possess various skills to be effective in practice including planning \citep{brohan2023can,huang2022language,logeswaran-etal-2022-shot}, reasoning \citep{wei2022chain,gao2023pal}, tool manipulation \citep{qin2023toolllm,patil2023gorilla,parisi2022talm,schick2023toolformer}, code generation \citep{sun2023adaplanner,logeswaran-etal-2022-shot}, among others.
In this work, we focus on building effective agents for web  \citep{yao2022react,zhou2023webarena} and embodied \citep{shridhar20alfworld} environments.

\textbf{Self-reflection from past experiences.}
An important capability for agents to succeed is the ability to learn from past experiences and update their behavior based on feedback.
Self-feedback \citep{madaan2023self,kim2023language,shinn2023reflexion} has emerged as an effective technique where a model inspects its own incorrect predictions, reflects on it to identify what went wrong and attempts to improve its prediction.
While self-feedback provides intra-task (i.e., per-episode) knowledge within a task based on immediate feedback, our approach offers an orthogonal and complementary aspect of inter-task knowledge (over multiple tasks) by considering multiple train tasks in offline data. %, offering a broader learning perspective.
AutoGuide enhances learning efficiency and credit assignment by utilizing detailed feedback from multiple tasks.
% , in contrast to methods that depend on immediate feedback from the previous episode.
% please note that when tasks change over episodes, feedback from the previous episode may not be relevant to the task in the next episode.
% Thus, AutoGuide allows for leveraging a richer learning signal, enabling the agent to quickly identify effective strategies and actions predictive of success by extracting beneficial knowledge across multiple tasks. As a result, AutoGuide speeds up the learning process and effectively addresses the credit assignment problem.
% Our approach shares similar motivations to self-refinement approaches, but unlike these methods which generate reflections on the fly, our approach first attempts to distill a set of guidelines from offline experiences and then retrieves the appropriate guidelines during inference for more accurate action prediction. 
However, these self-feedback approaches are complementary to \textsc{\Ours} and can be used in conjunction with our approach, as shown in our experiments (see \Cref{sec:main_result}).

\textbf{Leveraging natural language guidance.} Natural Language can be a rich source of information for agents to learn to act efficiently. Prior work has explored the notion of learning from human-written text manuals, which describe details about the environment \citep{branavan2012learning,hanjie2021grounding,zhong2020rtfm}.
Recent work has explored automatically generating such guidance in the form of chain-of-thought reasoning \citep{wei2022chain,yao2022react}, which emulates a thought process or rationale for agent's predictions.
In contrast to approaches which generate such guidance dynamically on the fly by imitating example guidance demonstrations provided by a human, our approach carefully compares trajectories in offline data to generate appropriate guidance and uses these guidelines for predicting better actions. 
ExpeL \citep{zhao2023expel} proposed a related approach to derive guidelines.
In contrast to ExpeL, where all guidelines are provided to an agent as a prompt, our guideline selection process is contextual, where guidelines relevant to the agent's current state are retrieved and used for prediction.
We show that this substantially improves over ExpeL's non-contextual guideline-based approach.

\def\trajsucc{\bm{\tau}^{i}_{+}}
\def\trajfail{\bm{\tau}^{i}_{-}}

\section{\textsc{\Ours}: Principled Method Based on Context-Aware Guidelines}\label{sec:method}
Our work is motivated by the increasing availability of offline experiences that agents or humans naturally accumulate through their interactions with the environment. \textsc{\Ours} aims to leverage this offline data to improve the decision-making of an LLM agent by generating helpful context-aware guidelines. This section details how \textsc{\Ours} automatically constructs these guidelines and applies them to guide action generation at test time.

%%%%%%%%%%%%%%%%%%%%%%%%%%%%%%%%%%%%%%%%%%%%%%%%%%%%%%%%%
\begin{figure}[t]
    \centering
    \includegraphics[width=\textwidth]{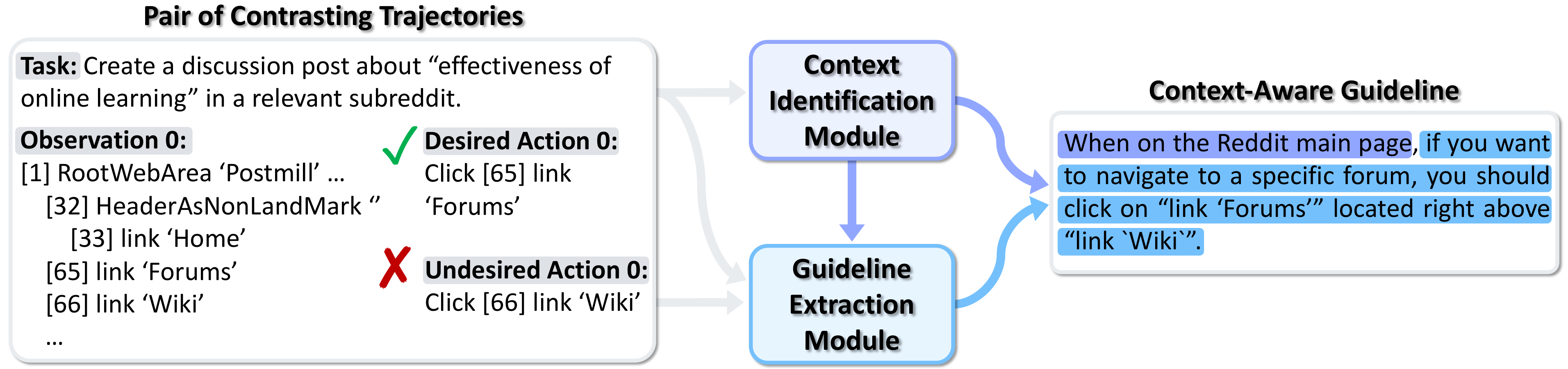}
    \caption{
    Context-aware guideline generation process based on a pair of contrastive trajectories $\trajsucc$ and $\trajfail$. In this example, the two trajectories start deviating from each other at $t\!=\!0$. The context identification module generates a description of the context at $t\!=\!0$ given $\bm{\tau}^{i}_{:0}$, and the guideline extraction module generates the corresponding guideline for that context.
    }
    \label{fig:guideline-method}
    \vspace{-0.5cm}
\end{figure}
%%%%%%%%%%%%%%%%%%%%%%%%%%%%%%%%%%%%%%%%%%%%%%%%%%%%%%%%%

\subsection{Problem Statement}\label{sec:problem-statement}
Formally, \textsc{\Ours} is given offline data $\mathcal{D}_{\mathrm{train}}\!=\!(\bm{\tau}^{1},...,\bm{\tau}^{N})$ that consist of $N$ trajectories from training tasks. Each trajectory $\bm{\tau}\!=\!(x_0,a_0,r_0,...,r_T)$ is a sequence of observations, actions, and rewards following the partially observable Markov decision process \cite{sutton1998rlbook}. The return of a trajectory is defined as the sum of rewards obtained throughout the trajectory: $R(\tau)\!=\!\smallsum_{t=0}^{T}{r_t}$.
The objective of \textsc{\Ours} is to distill knowledge from offline experiences into a useful natural language format, such that the extracted information helps to maximize the expected return $\mathbb{E}_{\tau}[R(\tau)]$ during test time.
% we actually maximize this: {$\mathbb{E}_{\tau \sim \mathcal{D}_{\mathrm{train}}}[R(\tau)]$}

\subsection{Extraction of Context-Aware Guidelines}\label{sec:extraction-context-aware-guidelines}
\textsc{\Ours} generates a set of context-aware guidelines by utilizing pairs of contrastive trajectories from offline data. Each context-aware guideline is expressed in concise natural language and follows a conditional structure, clearly describing the context in which the guideline is applicable.
Intuitively, contrasting a pair of trajectories with different returns provides important information about when and which actions are effective or ineffective in maximizing expected returns.
Building on this insight, we develop two modules for automatically extracting context-aware guidelines (see \Cref{fig:guideline-method}):

% \textbf{Context identification module.} \textcolor{blue}{The goal of this module is to describe the context of a trajectory. Specifically, let $\trajsucc$ and $\trajfail$ represent a contrasting pair of trajectories for the same task $i$ in offline data $\mathcal{D}_{\mathrm{train}}$, where $R(\trajsucc)\!>\!R(\trajfail)$. 
% By comparing these two trajectories, we find the deviation timestep $t$ at which they begin to diverge due to different actions. To generate the context corresponding to $t$, we construct a target trajectory $\bm{\tau}^{i}_{:t}\!:=\!(x_0,a_0,...,x_{t})$, which comprises a sequence up to $t$ from either trajectory and prompt an LLM based on our prompt template in \Cref{sec:appendix-prompt-template-context-summarization}:
% }
% \begin{align}\label{eqn:context-summarization-module}
% \begin{split}
% \textsc{context}\!\leftarrow\!\mathcal{M}_{\mathrm{context}}(\bm{\tau}^{i}_{:t}),
% \end{split}
% \end{align}
% For example, \Cref{fig:guideline-method} shows \textcolor{blue}{a pair of contrasting} trajectories related to the task $i$ of creating a discussion post about online learning in a relevant SubReddit. Given that these two trajectories have different actions at a deviating timestep $t\!=\!0$, our context identification module generates the following context from $\bm{\tau}^{i}_{:t}$ in \Cref{fig:guideline-method}: \emph{On the Reddit main page}. 

\textbf{Context identification module.} This module is responsible for abstracting the given partial trajectory into its \emph{context}, a concise natural language description of the agent's state.
% The goal of this module is to describe the context of a trajectory.
% In our setting, context refers to the status of the LLM agent, for example where it is or what it is doing.
More specifically, for a timestep $t$ and the corresponding trajectory $\bm{\tau}^{i}_{:t}\!:=\!(x_0,a_0,...,x_{t})$, we prompt LLMs to clearly describe the agent's status:
\begin{align}\label{eqn:context-summarization-module}
\begin{split}
\textsc{context}\!\leftarrow\!\mathcal{M}_{\mathrm{context}}(\bm{\tau}^{i}_{:t}),
\end{split}
\end{align}
Our prompt templates for the context identification module are shown in \Cref{sec:appendix-prompt-template-context-summarization}.

%%%%%%%%%%%%%%%%%%%%%%%%%%%%%%%%%%%%%%%%%%%%%%%%%%%%%%%%%
\begin{figure}[t]
\begin{minipage}[t]{0.47\textwidth}
\begin{algorithm}[H]
   \small
   \caption{Extracting context-aware guidelines from offline data}
   \label{alg:ours_train}
   \begin{algorithmic}
      \STATE {\bfseries Input:} Offline data $\mathcal{D}_{\mathrm{train}}$, context identification module $\mathcal{M}_{\mathrm{context}}$, guideline extraction module $\mathcal{M}_{\mathrm{guideline}}$
      \STATE Initialize context-aware guideline dictionary $\mathcal{G}$
      \FOR{Each pair $\trajsucc, \trajfail \in \mathcal{D}_{\mathrm{train}}$ }
           \STATE \textcolor[RGB]{34, 108, 192}{\# Identify the context from a trajectory}
           \STATE Find the deviating timestep $t$ from $\trajsucc$ and $\trajfail$
           \STATE $\textsc{context} \gets \mathcal{M}_{\mathrm{context}}(\bm{\tau}^{i}_{:t})$
           \STATE \textcolor[RGB]{34, 108, 192}{\# Check if the current context matches any\\\hspace{0.33cm}existing contexts}
           \IF{$\textsc{context}\notin\mathcal{G}$}
               \STATE $\mathcal{G}[\textsc{context}] = \{\}$
           \ENDIF
           \STATE \textcolor[RGB]{34, 108, 192}{\# Generate the context-aware guideline}
           \STATE $\textsc{guideline}\!\!\gets\!\! \mathcal{M}_{\mathrm{guideline}}(\trajsucc, \trajfail, \textsc{context})$
           \STATE $\mathcal{G}[\textsc{context}]\!\!\gets\!\!\mathcal{G}[\textsc{context}]\cup\{\textsc{guideline}\}$
      \ENDFOR
      \STATE \textbf{Return} Context-aware guideline dictionary $\mathcal{G}$
      \vspace{0.24cm}
   \end{algorithmic}
\end{algorithm}
\end{minipage}
\hfill
% \begin{minipage}[t]{0.53\textwidth}
\begin{minipage}[t]{0.50\textwidth}
\begin{algorithm}[H]
   \small
   \caption{Applying context-aware guidelines at test time}
   \label{alg:ours_test}
   \begin{algorithmic}
      \STATE {\bfseries Input:} Context-aware guideline dictionary $\mathcal{G}$, context identification module $\mathcal{M}_{\mathrm{context}}$, guideline selection module $\mathcal{M}_{\mathrm{select}}$, LLM agent policy $\pi$
      % \STATE {\bfseries Parameter:} Horizon $T$, max number of guidelines $k$
      \STATE Initialize test trajectory $\bm{\tau}=\{x_0\}$
      \FOR{Each timestep $t$}
           \STATE \textcolor[RGB]{34, 108, 192}{\# Identify the current context from a trajectory}
           \STATE $\textsc{context} \gets \mathcal{M}_{\mathrm{context}}(\bm{\tau})$
           \STATE \textcolor[RGB]{34, 108, 192}{\# If the current context matches any existing\\\hspace{0.26cm} ones, perform top-$k$ guideline selection}
           \IF{$\textsc{context}\in\mathcal{G}$}
           \STATE $\textsc{guidelines}\gets\mathcal{M}_{\mathrm{select}}(\textsc{context},\bm{\tau} ; \mathcal{G}, k)$
           \ELSE
           \STATE $\textsc{guidelines}\gets\varnothing$
           \ENDIF
           \STATE \textcolor[RGB]{34, 108, 192}{ \# Action selection based on guidelines}
           \STATE $a_t \sim \pi(\bm{\tau}, \textsc{context},\textsc{guidelines})$ 
           \STATE Execute action $a_t$ and observe $x_{t+1}$
           \STATE Update trajectory $\bm{\tau} \gets \bm{\tau} \cup \{\textsc{context},a_t,x_{t+1}\}$
       \ENDFOR
   \end{algorithmic}
\end{algorithm}
\end{minipage}
\vspace{-0.5cm}
\end{figure}
%%%%%%%%%%%%%%%%%%%%%%%%%%%%%%%%%%%%%%%%%%%%%%%%%%%%%%%%%

% \textbf{Guideline extraction module.} This module aims to generate a desired guideline corresponding to the generated context. We extract a useful natural language guideline by examining \textcolor{blue}{the paired contrastive trajectories $\trajsucc$ and $\trajfail$ with respect to the context: }
% \begin{align}\label{eqn:action-extraction-module}
% \begin{split}
% \text{\textsc{guideline}}\!\leftarrow\! \mathcal{M}_{\mathrm{guideline}}(\trajsucc, \trajfail,\textsc{context}),
% \end{split}
% \end{align}
% where we refer to \Cref{sec:appendix-prompt-template-guideline-extraction} for our prompt template. As an example, this module generates the following context-aware guideline for the context described in \Cref{fig:guideline-method}:  \emph{When on the Reddit main page, if you want to navigate to a specific forum, you should click on ``link `Forums'\hspace{-0.02cm}'' located right above ``link `Wiki'\hspace{-0.02cm}''}.

\textbf{Guideline extraction module.}
This module aims to generate a desired guideline corresponding to a specific context.
Let $\trajsucc$ and $\trajfail$ represent a contrasting pair of trajectories for the same task $i$ in offline data $\mathcal{D}_{\mathrm{train}}$, where $R(\trajsucc)\!>\!R(\trajfail)$.
We want to contrast the pair of trajectories to find desired behaviors at an important timestep.
To do this, we compare these two trajectories to find the deviation timestep $t$ at which they begin to diverge due to different actions. 
Then we apply the context identification module to summarize the context for the shared part of the trajectory $\bm{\tau}^{i}_{:t}$.
Eventually, we extract a useful natural language guideline by examining the paired contrastive trajectories $\trajsucc$ and $\trajfail$ with respect to the context:
\begin{align}\label{eqn:action-extraction-module}
\begin{split}
\text{\textsc{guideline}}\!\leftarrow\! \mathcal{M}_{\mathrm{guideline}}(\trajsucc, \trajfail,\textsc{context}),
\end{split}
\end{align}
where we refer to \Cref{sec:appendix-prompt-template-guideline-extraction} for our prompt template. 
As an example, the paired trajectories in \Cref{fig:guideline-method} deviate from timestep $t=0$, for which the context is summarized as \emph{On the Reddit main page}. 
This module then generates the following context-aware guideline: \emph{When on the Reddit main page, if you want to navigate to a specific forum, you should click on ``link `Forums'\hspace{-0.02cm}'' located right above ``link `Wiki'\hspace{-0.02cm}''}.
% For example, \Cref{fig:guideline-method} shows \textcolor{blue}{a pair of contrasting} trajectories related to the task $i$ of creating a discussion post about online learning in a relevant SubReddit. Given that these two trajectories have different actions at a deviating timestep $t\!=\!0$, our context identification module generates the following context from $\bm{\tau}^{i}_{:t}$ in \Cref{fig:guideline-method}: \emph{On the Reddit main page}. 

\textbf{Construction of context-aware guidelines.} 
We collect context-aware guidelines $\mathcal{G}$ by iterating through available pairs in the paired offline data and organize the guidelines in a dictionary format, using the context as the key and the corresponding guidelines as the value (see Algorithm \ref{alg:ours_train}). In particular, we observe that the context identification module occasionally produces contexts that describe the same situation but are expressed slightly differently. To minimize redundancy, we employ an LLM to determine if the current context corresponds to any previously identified context. If a match is found, we reuse the existing context; otherwise, we introduce a new context into our dictionary $\mathcal{G}$. The specific prompt template for this context-matching procedure is in \Cref{sec:appendix-prompt-template-context-matching}.

\subsection{Applying Context-Aware Guidelines at Test Time}\label{sec:applying-context-aware-guidelines}
After extracting a set of context-aware guidelines $\mathcal{G}$ from offline experiences, our method employs these guidelines to enhance the decision-making of an LLM agent during testing. 
At each timestep, \textsc{\Ours} identifies the $\textsc{context}$ of the current test trajectory $\bm{\tau}$ up to timestep $t$ (which represents the agent's interactions up to the current time-step) using our context identification module $\mathcal{M}_\mathrm{context}$. 
% Then, we employ the LLM designed for identifying a matching context in \Cref{sec:extraction-context-aware-guidelines} to find a match and retrieve its corresponding guidelines.
Our guideline selection module $\mathcal{M}_{\mathrm{select}}$ then selects relevant guidelines for $\textsc{context}$ from $\mathcal{G}$. More specifically, the module applies $\textsc{context}$ as the key to fetch a set of possible guidelines $\mathcal{G}[\textsc{context}]$. If there are more than $k$ guidelines in $\mathcal{G}[\textsc{context}]$, $\mathcal{M}_{\mathrm{select}}$ prompts an LLM to choose top-$k$ guidelines for the specific $\bm{\tau}$:
% \begin{align}\label{eqn:guideline-selection-module}
% \begin{split}
% \mathcal{M}_{\mathrm{select}}(\textsc{context},\mathcal{G}[\textsc{context}],\bm{\tau}),
% \end{split}
% \end{align}
\begin{align}\label{eqn:guideline-selection-module}
\begin{split}
\textsc{relevant guidelines} \gets \mathcal{M}_{\mathrm{select}}(\textsc{context}, \bm{\tau}; \mathcal{G},k),
\end{split}
\end{align}
where \Cref{sec:appendix-prompt-template-guideline-retrieval} details the prompt template for this selection procedure. 
Subsequently, \textsc{\Ours} incorporates both the context and relevant guidelines into the agent's action generation prompt. Therefore, the agent selects an action by considering the provided context and guidelines (see \Cref{fig:guideline-intro} for an example). This process iterates until the end of the test trajectory (see \Cref{alg:ours_test}).

\textbf{Key benefits of \textsc{\Ours}}. First, the extraction of context-aware guidelines in \textsc{\Ours} offers the inherent benefit of providing relevant guidelines for the context of interest. This capability is important since neglecting the specific context in which a guideline applies can confuse the agent's decision-making process. The second key benefit is the generation of concise natural language guidelines, which can be seamlessly incorporated into any prompt-based LLM agent. Lastly, \textsc{\Ours} generates guidelines at the individual context level rather than at the trajectory level. Given that a single incorrect action can lead to a complete failure, it is essential to provide detailed assistance in each action selection process. With these advantages, we demonstrate in the next section that our approach significantly enhances the performance of LLM agents.
%%%%%%%%%%%%%%%%%%%%%%%%%%%%%%%%%%%%%%%%%%%%%%%%%%%%%%%%%
\begin{figure}[!t]
    \centering
    \includegraphics[width=0.9\textwidth]{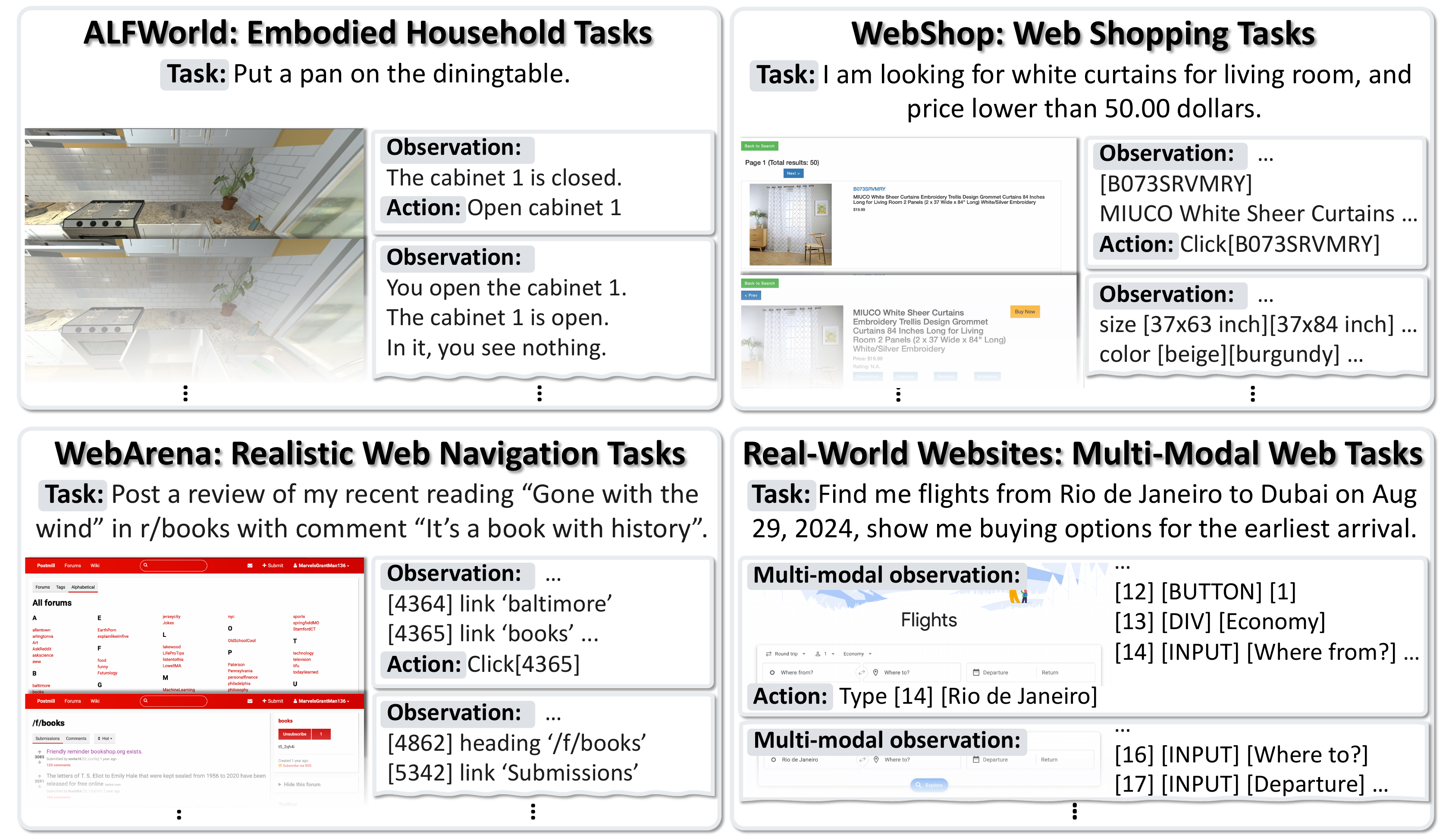}
    \vspace{-0.1 cm}
    \caption{Sequential decision-making benchmark domains considered in our work: ALFWorld \citep{shridhar20alfworld}, WebShop \citep{yao2022webshop}, WebArena \citep{zhou2023webarena}, and multi-modal real-world websites. Graphic credit: \cite{shridhar20alfred,yao2022webshop,zhou2023webarena,gtravel}.}
    \label{fig:domain}
    % \vspace{-0.8cm}
\end{figure}
%%%%%%%%%%%%%%%%%%%%%%%%%%%%%%%%%%%%%%%%%%%%%%%%%%%%%%%%%

\section{Evaluation}\label{sec:evaluation}
This section demonstrates the efficacy of \textsc{\Ours} by conducting experiments on a diverse suite of sequential decision-making benchmark domains. We also perform important analyses about \textsc{\Ours}, such as the ablation study of different \textsc{\Ours} components, comparison to in-context learning, and generalization to out-of-domain tasks.
We refer to \Cref{sec:appendix-evaluation-details} for additional experimental details.

\subsection{Evaluation Setup}\label{sec:evaluation-setup}
\subsubsection{Sequential Decision-Making Benchmark Domains}
We consider the following interactive sequential decision-making benchmarks to study various aspects of \textsc{\Ours} (see \Cref{fig:domain}):
\begin{itemize}[leftmargin=*, wide, labelindent=0pt, topsep=0pt]
    \itemsep 0pt 
    \item \textbf{ALFWorld \cite{shridhar20alfworld}:} In this embodied benchmark, an LLM agent interacts with an environment to carry out household tasks, such as placing a pan on the dining table. Observations and actions are expressed in natural language statements, and the agent must navigate through the space and manipulate objects to successfully complete the tasks. 
    \item \textbf{WebShop \cite{yao2022webshop}:} This interactive web environment simulates the task of online shopping on an e-commerce website. The agent’s goal is to understand a text instruction and buy a product that meets specified criteria. This involves querying the website’s search engine, understanding the descriptions and details of each item, and selecting necessary options. 
    \item \textbf{WebArena \cite{zhou2023webarena}:} This web-based benchmark introduces realistic environments by replicating the functionality and data found in popular web domains (e.g., Gitlab, Reddit, Wikipedia). Compared to WebShop, WebArena presents more challenges and difficulties for an LLM agent due to its large observation and action space, along with tasks that involve longer planning horizons. We focus oe the Reddit domain for the main WebArena experiments.
    \item \textbf{Real-world multi-modal websites:} Finally, we consider evaluating \textsc{\Ours} on a variety of real-world website tasks. These span from a collaborative software development platform (e.g., GitHub) to a flight search engine (e.g., Google Flights) and an online education platform (e.g., Coursera). Please refer to \Cref{sec:appendix_real_website_details} for example tasks. In particular, in comparison to WebShop and WebArena, we design our tasks to be multi-modal such that the agent must consider both visual (e.g., images) and textual information (e.g., HTML) to complete these tasks.
\end{itemize}

%%%%%%%%%%%%%%%%%%%%%%%%%%%%%%%%%%%%%%%%%%%%%%%%%%%%%%%%%
\begin{figure}[t!]
    \centering
    \includegraphics[width=\textwidth]{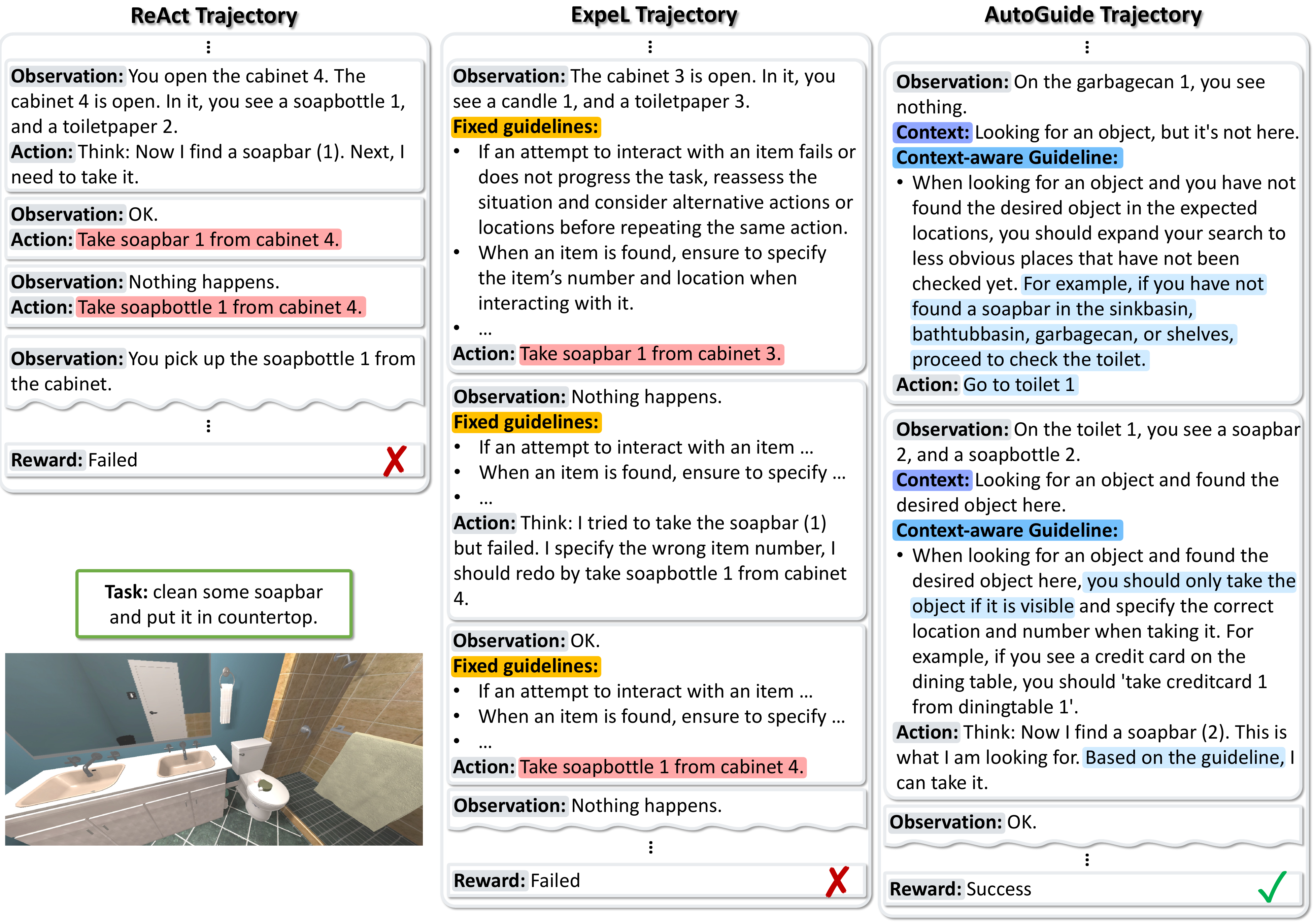}
    \vspace{-0.4cm}
    \caption{Trajectories of ReAct, ExpeL, and \textsc{\Ours} from the same test task. ReAct (Left) chose the wrong item, consequently failing the task in the end. ExpeL (middle) was confused by guidelines that were irrelevant to current context, leading to incorrect reasoning and actions. \textsc{\Ours} (right) selects relevant guidelines to the agent's context, enabling the agent to accomplish the task.}
    \label{fig:guideline-example}
    \vspace{-0.3cm}
\end{figure}
%%%%%%%%%%%%%%%%%%%%%%%%%%%%%%%%%%%%%%%%%%%%%%%%%%%%%%%%%

\subsubsection{Baselines}
We compare \textsc{\Ours} against the following baseline approaches to study the effect of context-aware guidelines (refer to \Cref{sec:appendix-evaluation-details} for more details):
\begin{itemize}[leftmargin=*, wide, labelindent=0pt, topsep=0pt]
    \itemsep 0pt 
    \item \textbf{ReAct \cite{yao2022react}:} This LLM-based planning method integrates reasoning and acting to address sequential decision-making tasks. However, it does not leverage offline experiences and thus suffers from the limited understanding of pre-trained LLMs in downstream domains. 
    \item \textbf{ExpeL \cite{zhao2023expel}}: This method also extracts natural language knowledge from offline data. However, it fails to consider the applicability of guidelines and does not generate context-aware guidelines. Instead, it provides all guidelines to an LLM agent without filtering out irrelevant ones based on the current context. ExpeL has two contributions, the guideline generation and in-context example selection module. Because the latter is orthogonal to our analysis and can be seamlessly combined with our method, we consider ExpeL with guidelines in our experiments.
    \item \textbf{Reflexion \cite{shinn2023reflexion}:} This approach converts environmental feedback into text statements to assist an LLM agent (e.g., ReAct) in the next trial. The baseline generates valuable feedback about solving a specific test task. We demonstrate how context-aware guidelines derived by \textsc{\Ours} can be combined with the feedback.
\end{itemize}

\subsubsection{Implementation}
We collect offline experiences either by running ReAct and Reflexion, or incorporating human demonstrations.
We use ReAct with GPT-3.5-turbo as our base LLM agent for WebShop and ALFWorld and GPT-4-turbo for WebArena. 
For each benchmark, we apply the same GPT model for action generation, context identification, and guideline selection. We extract context-aware guidelines from offline data with GPT-4-turbo and evaluate their effectiveness by applying them to the test set with non-overlapping tasks. 
We refer to \Cref{sec:appendix-evaluation-details} for more details.

%%%%%%%%%%%%%%%%%%%%%%%%%%%%%%%%%%%%%%%%%%%%%%%%%%%%%%%%%
\begin{table}[t]
    \centering
    \small
    \renewrobustcmd{\bfseries}{\fontseries{b}\selectfont}
    \sisetup{detect-weight,mode=text,group-minimum-digits=4}
    {
        \tabcolsep=0.1cm
        \begin{tabular}[t]{l c c c c c c c}
	\toprule
	\multirow{2}{*}[-2pt]{Algorithm} & & \multirow{2}{*}[-2pt]{\makecell{Offline\\data?}} & \multirow{2}{*}[-2pt]{\makecell{Context\\aware?}} & \multicolumn{1}{c}{ALFWorld \cite{shridhar20alfworld}} & \multicolumn{2}{c}{WebShop \cite{yao2022webshop}} & \multicolumn{1}{c}{WebArena \cite{zhou2023webarena}}\\
		\cmidrule(lr){5-5} \cmidrule(lr){6-7} \cmidrule(lr){8-8} & & & & Success Rate (SR)$\uparrow$ & Reward$\uparrow$ & SR$\uparrow$ & SR$\uparrow$\\ \midrule
		ReAct \cite{yao2022react} & & \myredmark & \myredmark & 54.5\text{\%} & 66.4 & 30\text{\%}& 8.0\text{\%}\\ 
		ExpeL \cite{zhao2023expel} & & \mygreencheck & \myredmark & 59.0\text{\%} & 60.9 & 35\text{\%} & 21.8\text{\%}\\ 
		\textsc{\Ours} & & \mygreencheck & \mygreencheck & \bfseries 79.1\text{\%} & \bfseries 73.4 & \bfseries 46\text{\%} & \bfseries 47.1\text{\%}\\ \midrule
		ReAct \cite{yao2022react} &+ Reflexion \cite{shinn2023reflexion} & \myredmark & \myredmark & 67.2\text{\%} & 77.1 & 51\text{\%} & N/A\\ 
		ExpeL \cite{zhao2023expel} &+ Reflexion \cite{shinn2023reflexion} & \mygreencheck & \myredmark & 71.6\text{\%} & 71.7 & 42\text{\%} & N/A\\ 
		\textsc{\Ours} &+ Reflexion \cite{shinn2023reflexion} & \mygreencheck & \mygreencheck & \bfseries 88.1\text{\%} & \bfseries 81.4 & \bfseries 57\text{\%} & N/A\\
            \bottomrule
	\end{tabular}
    }
    \vspace{5pt}
    \caption{Test reward and success rate on ALFWorld, WebShop, and WebArena. The base agent model for ALFWorld and WebShop is GPT-3.5-turbo and for WebArena is GPT-4-turbo. Reflexion is done by GPT-4-turbo for at most 3 trials. In our experiments, due to token limit of GPT, we did not experiment with Reflexion on WebArena tasks.}
    \vspace{-0.6cm}
    \label{table:main-result} 
\end{table}
%%%%%%%%%%%%%%%%%%%%%%%%%%%%%%%%%%%%%%%%%%%%%%%%%%%%%%%%%

\subsection{Main Results}\label{sec:main_result}

\textbf{Q1.} \textit{How effective is \textsc{\Ours} compared to baselines without context-aware guidelines?}

To answer this question, we compare methods on ALFWorld, WebShop, and WebArena benchmarks. The performance on the test datasets is presented in \Cref{table:main-result}. There are three notable observations:
\begin{enumerate}[leftmargin=*, wide, labelindent=0pt, topsep=0pt]
    \itemsep 0pt 
    \item \textbf{Effectiveness of context-aware guidelines.} Our approach surpasses baseline performance in both ALFWorld and WebShop, achieving the highest test rewards and success rates in \Cref{table:main-result}. These results highlight the effectiveness of employing context-aware guidelines in language-based decision-making domains. 
    To further examine the action selection differences among ReAct, ExpeL, and our method, we present their trajectories in \Cref{fig:guideline-example}. We observed that ReAct makes common mistakes such as trying to take soapbar that is not visible, or taking a soapbottle instead of soapbar due to their similar names. Both ExpeL and \textsc{\Ours} improve on this by extracting guidelines from similar mistakes in the offline experience. However, ExpeL often erroneously applies incorrect guidelines due to the availability of all guidelines at each timestep. In \Cref{fig:guideline-example}, ExpeL mistakenly attends to the second guideline \emph{``ensure to specify the item's number and location ...''}, leading to wrong reasoning and action. \textsc{\Ours} presents relevant guidelines at necessary moments, enabling accurate task completion by avoiding the mistakes seen in ExpeL and ReAct.
    \item \textbf{Importance of providing pertinent knowledge.} ExpeL approach helps ReAct by extracting knowledge from offline experiences, but its impact is not as significant as \textsc{\Ours}. Recall that for ExpeL, the guidelines are neither generated for specific contexts at training time nor selected to only provide context-aware guidelines at test time. As a result, irrelevant guidelines can be introduced to an agent, potentially causing confusion for the agent. Consequently, the result highlights the significance of providing relevant guidelines conditioned on contexts for LLM agents.
    \item \textbf{Scalability to complex environments.} We conduct experiments on WebArena-Reddit, which features more diverse tasks on realistic and complex websites requiring longer action sequences. This domain has a larger observation space and a more complex action space (e.g., scrolling). \Cref{table:main-result} presents the results, where \textsc{\Ours} achieves the highest success rate with a significant margin when compared to ReAct and ExpeL. We observe that ReAct scores low task success rate (8.0\%) in WebArena due to the complex observation and action spaces and longer task horizon. In ExpeL, the issue of presenting all guidelines to an agent is exacerbated in the WebArena compared to simpler environments like ALFWorld and WebShop. WebArena's wide variety of tasks across different domains requires a larger number of guidelines to cover the knowledge needed for all tasks and domains. This results in either an overload of irrelevant guidelines that could mislead the agent or a lack of crucial information when the number of guidelines is limited, as suggested in ExpeL~\cite{zhao2023expel}. In contrast, \textsc{\Ours} achieves a more significant performance enhancement (47.1\%) compared to ExpeL (21.8\%) by efficiently providing pertinent guidelines and minimizing the burden on context capacity. We refer to \Cref{fig:state-and-guideline-webarena} for a list of example contexts and guidelines.
\end{enumerate}

%%%%%%%%%%%%%%%%%%%%%%%%%%%%%%%%%%%%%%%%%%%%%%%%%%%%%%%%%
\begin{table}[t]
\small
\centering
\begin{minipage}[t]{.38\textwidth}
    % \vspace{1.1cm}
    \centering
    \renewrobustcmd{\bfseries}{\fontseries{b}\selectfont}
    \sisetup{detect-weight,mode=text,group-minimum-digits=4}
    {
        \tabcolsep=0.04cm
        \begin{tabular}[t]{cccccc}
		\toprule
		\multirow{2}{*}[-2pt]{Algorithm} & & \multicolumn{1}{c}{GitHub} & \multicolumn{1}{c}{Flights} & \multicolumn{1}{c}{Coursera}\\
		\cmidrule(lr){3-3} \cmidrule(lr){4-4} \cmidrule(lr){5-5} 
            & & SR$\uparrow$ & SR$\uparrow$ & SR$\uparrow$ & \\ \midrule
		SoM \cite{yang2023setofmark} & & 2/30 & 5/20 & 1/20\\ 
		\textsc{\Ours} & & \bfseries 19/30 & \bfseries 9/20 & \bfseries 14/20\\
            \bottomrule
	\end{tabular}
    }
    \vspace{5pt}
    \caption{Test results of \textsc{\Ours} on 3 real-world web domains within multi-modal settings. The base agent model runs with GPT-4V and applying context-aware guidelines significantly improves the performance.}
    \label{table:real-website-result} 
\end{minipage}
\hfill
\begin{minipage}[t]{.33\textwidth}
    \centering
    \renewrobustcmd{\bfseries}{\fontseries{b}\selectfont}
    \sisetup{detect-weight,mode=text,group-minimum-digits=4}
    {
        \tabcolsep=0.04cm
        \begin{tabular}[t]{cccc}
		\toprule
		\multirow{2}{*}[-2pt]{Algorithm} & & \multicolumn{2}{c}{WebShop} \\
		\cmidrule(lr){3-4} 
            & & Reward$\uparrow$ & SR$\uparrow$ \\ \midrule
		ReAct (1-shot) & & 66.4 & 30\text{\%} \\ 
		ReAct (2-shot) & & 66.0 & 35\text{\%} \\ 
		ReAct (4-shot) & & 70.2 & 37\text{\%} \\ 
		ReAct (6-shot) & & 71.0 & 38\text{\%} \\ 
		\textsc{\Ours} & & \bfseries 73.4 & \bfseries 46\text{\%} \\
            \bottomrule
	\end{tabular}
    }
    \vspace{5pt}
    \caption{Analysis of \textsc{\Ours} against ReAct with varying numbers of in-context examples.}
    \label{table:in-context-result} 
\end{minipage}
\hfill
\begin{minipage}[t]{.24\textwidth}
    \centering
    \renewrobustcmd{\bfseries}{\fontseries{b}\selectfont}
    \sisetup{detect-weight,mode=text,group-minimum-digits=4}
    {
        \tabcolsep=0.04cm
        \begin{tabular}[t]{cccc}
		\toprule
		\multirow{2}{*}[-2pt]{Top-$k$} & & \multicolumn{1}{c}{WebShop} \\
		\cmidrule(lr){3-3} 
            & & SR$\uparrow$ \\ \midrule
		$k\!=\!0$ & & 30\text{\%} \\ 
		$k\!=\!1$ & & 42\text{\%} \\ 
		$k\!=\!2$ & & 46\text{\%} \\ 
		$k\!=\!3$ & & 47\text{\%} \\ 
		$k\!=\!5$ & & 43\text{\%} \\ 
            \bottomrule
	\end{tabular}
    }
    \vspace{5pt}
    \caption{Ablation study of \textsc{\Ours} using various top-$k$ values.}
    \label{table:top-k-result} 
\end{minipage}
\vspace{-1cm}
\end{table}
% HL: How about providing more examples(higher k) for top-k? if the performance further drops with larger k, this implies that just providing more examples is not an effective strategy, and it also supports why our method should perform better than ExpeL.

%%%%%%%%%%%%%%%%%%%%%%%%%%%%%%%%%%%%%%%%%%%%%%%%%%%%%%%%%

\textbf{Q2.} \textit{How does \textsc{\Ours} perform when combined with test-time self-feedback approaches?}

Our context-aware guidelines effectively provide \textit{inter-task} knowledge by considering multiple tasks in offline data. Meanwhile, self-feedback methods (e.g., Reflexion) offer \textit{intra-task} knowledge based on environmental feedback during test time. In this question, we explore the effectiveness of integrating both inter-task and intra-task information. The results presented in \Cref{table:main-result} demonstrate that the combination of \textsc{\Ours} with Reflexion achieves the highest performance in the WebShop and ALFWorld benchmarks. Hence, we find that our context-aware guidelines positively complement the intra-task knowledge of Reflexion. Another observation from \Cref{table:main-result} is that, while ExpeL + Reflexion outperforms ExpeL alone, this combination is not as effective as other approaches. This limitation may stem from ExpeL introducing irrelevant knowledge, potentially leading to conflicts with Reflexion's feedback and having an adverse impact on the decision-making process.

\textbf{Q3.} \textit{Can \textsc{\Ours} generate context-aware guidelines for multi-modal inputs?}

Going beyond text-only inputs is an essential step toward building capable agents for solving real-world environments and tasks.
We test \textsc{\Ours} in a complex multi-modal setting, where each observation includes image and text information. Specifically, we introduce a set of real-world website navigation tasks in 3 domains: GitHub, Google Flights, and Coursera. For these multi-modal tasks, we employ the Set-of-Marks (SoM) agent \citep{yang2023setofmark,koh2024visualwebarena} as our base method.
The SoM prompting improves the visual grounding capabilities of large multi-modal models such as GPT-4V by adding visually distinguishable marks to image inputs \citep{yang2023setofmark}. We apply \textsc{\Ours} with GPT-4V to generate natural language context-aware guidelines from collected trajectories with both image and text observations. \Cref{table:real-website-result} shows the effectiveness of \textsc{\Ours}, demonstrating its generalization ability to complex real-world multi-modal settings. We refer to \Cref{fig:state-and-guideline-real} for example context-aware guidelines.
% We then apply GPT-4V to generate natural language context-aware guidelines from collected trajectories with both image and text observations. 
%We run SoM\cite{} as our base VLM agent on this set of real-world website navigation tasks, which blablabla and does not consider offline data.
% Our implementation of \Ours{} in this setting takes images and texts as input and generate natural language guidelines that also consider visual knowledge (e.g. ).
% \textcolor{red}{todo. please briefly explain about som and explain that som does not consider offline data, similar to React, and we use this baseline for our multi-modal experiments in Question 3 for real-world web navigation experiments.; please prepare some examples for the appendix.}

\subsection{Analyses of \textsc{\Ours}}\label{sec:analysis}
\textbf{Q4.} \textit{How does \textsc{\Ours} compare to ReAct with varying numbers of in-context examples?}

\Cref{table:in-context-result} shows that, while increasing the number of in-context examples for ReAct gradually improves performance, there is a plateau at a certain number of shots. Additionally, ReAct with more than 6 shots often exceeds the token limit of GPT-3.5-turbo. These results indicate that directly inputting raw trajectories into ReAct for in-context learning is not an effective way to fully leverage offline data. In contrast, \textsc{\Ours} extracts knowledge from entire training trajectories by summarizing them into concise context-aware guidelines, making them easy to integrate with prompt-based agents.

\textbf{Q5.} \textit{How does altering the number of top-$k$ guidelines impact the performance of \textsc{\Ours}?}

We conducted an ablation study on WebShop using various values of $k$ in \Cref{table:top-k-result}. We find that employing context-aware guidelines consistently outperforms the no-guideline baseline ($k\!=\!0$; ReAct). The $k=3$ yields the best performance. The largest $k$ value of 5 can lead an LLM agent to overthink, potentially resulting in a slight decrease in performance. Conversely, a smaller $k$, like $k\!=\!1$, may cause LLM to overlook additional helpful guidelines, leading to slightly worse performance.

\textbf{Q6.} \textit{How do \textsc{\Ours}'s context-aware guidelines generalize to out-of-domain environments?}

We conduct an experiment to further demonstrate \textsc{\Ours}’s out-of-domain capability across different domains but relevant tasks. We extract context-aware guidelines from WebShop and apply them to WebArena-Shopping, which is a distinct domain with variations in observation/action spaces, task intentions, and episodic horizons. For this domain adaptation case, we additionally incorporate a grounding module to align the context-aware guidelines from WebShop to WebArena’s observations based on GPT-4-Turbo. As shown in \Cref{table:out-of-domain-result}, the transferred guidelines bring a notable improvement in success rates compared to the ReAct baseline in WebArena Shopping.

\textbf{Q7.} \textit{How does each component of \textsc{\Ours} contribute to the final results?}

We evaluate the impact of different components within \textsc{\Ours} on its performance in WebShop, as detailed in \Cref{table:analysis-ablation}. We examine two variants: ReAct+CI and ReAct+GES. 
The ReAct+CI, which incorporates contexts into observations without guidelines, shows improvement over ReAct. This suggests that contexts enhance decision-making by verifying the current state before action selection. 
ReAct+GES, which generates guidelines from trajectories without contexts and employs GPT-3.5-turbo for guideline selection, also enhances performance but is less effective than the full \textsc{\Ours}. 
This indicates that choosing relevant guidelines based on the trajectory alone is more challenging than using contexts. Therefore, integrating both context summaries and guidelines is crucial for maximizing the benefits of \textsc{\Ours}.

%%%%%%%%%%%%%%%%%%%%%%%%%%%%%%%%%%%%%%%%%%%%%%%%%%%%%%%%%
\begin{table}[t]
\small
\centering
\begin{minipage}{.45\textwidth}
    \vspace{0.2cm}
    \centering
    \renewrobustcmd{\bfseries}{\fontseries{b}\selectfont}
    \sisetup{detect-weight,mode=text,group-minimum-digits=4}
    {
        \tabcolsep=0.04cm
        \begin{tabular}[t]{cccc}
		\toprule
		\multirow{1}{*}[-2pt]{Algorithm} & & \multicolumn{1}{c}{WebArena--Shopping} \\
		\cmidrule(lr){3-3} 
            & & SR$\uparrow$ \\ \midrule
		ReAct & & 10.2\text{\%} \\ 
		\textsc{\Ours} & & 20.4\text{\%} \\ 
            \bottomrule
	\end{tabular}
    }
    \vspace{5pt}
    \caption{Out-of-distribution generalization of context-aware guidelines from WebShop on the 98 WebArena--Shopping tasks that have a product in the intent template.}
    \label{table:out-of-domain-result} 
\end{minipage}
\hfill
\begin{minipage}{.52\textwidth}
    \centering
    \renewrobustcmd{\bfseries}{\fontseries{b}\selectfont}%enable bfseries in siunitx tables
    \sisetup{detect-weight,mode=text,group-minimum-digits = 4}
    {
        \tabcolsep=0.15cm
        \begin{tabular}[t]{cccc}
        \toprule
            Algorithm&CI&GES& \text{WebShop SR$\uparrow$} \\ \midrule
		ReAct & \myredmark&\myredmark& 30\text{\%} \\
            %ExpeL & \myredmark&\mygreencheck&\myredmark& 35\text{\%} \\
		 ReAct + CI  &  \mygreencheck&\myredmark & 36\text{\%} \\
	   ReAct + GES & \myredmark&\mygreencheck & 37\text{\%} \\
		\textsc{\Ours}& \mygreencheck&\mygreencheck& \bfseries 46\text{\%} \\
            \bottomrule
	\end{tabular}
    }
    \vspace{5pt}
    \caption{Ablation study of \textsc{\Ours}, analyzing each module's contribution in WebShop. CI denotes our context identification module, and GES denotes the guideline extraction and selection modules.}
    \label{table:analysis-ablation} 
\end{minipage}
\vspace{-0.9cm}
\end{table}
%%%%%%%%%%%%%%%%%%%%%%%%%%%%%%%%%%%%%%%%%%%%%%%%%%%%%%%%%
\section{Conclusion}\label{sec:conclusion}
We present \textsc{\Ours}, an effective framework for exploiting important domain knowledge from offline experiences for improving decision-making with pre-trained LLMs. We proposed to generate context-aware guidelines that can be incorporated into prompts for LLM agents. As \textsc{\Ours} extracts the guidelines by contrasting trajectories in offline data, the resulting context-aware guidelines carry critical information for preventing failures in the domains. For inference, it provides the guidelines pertinent to each of the different context that LLM agents encounter, which can make pre-trained LLMs strong decision-making agents in the downstream domains. Empirically, we showed that \textsc{\Ours} outperforms strong baselines by a large margin and achieves outstanding performance in decision-making benchmarks.

\section{Acknowledgements}
This work was supported in part by LG AI Research.

\bibliographystyle{unsrt}
\bibliography{main}

\begin{thebibliography}{10}

\bibitem{wang2023survey}
Lei Wang, Chen Ma, Xueyang Feng, Zeyu Zhang, Hao Yang, Jingsen Zhang, Zhiyuan Chen, Jiakai Tang, Xu~Chen, Yankai Lin, et~al.
\newblock A survey on large language model based autonomous agents.
\newblock {\em arXiv preprint arXiv:2308.11432}, 2023.

\bibitem{xi2023rise}
Zhiheng Xi, Wenxiang Chen, Xin Guo, Wei He, Yiwen Ding, Boyang Hong, Ming Zhang, Junzhe Wang, Senjie Jin, Enyu Zhou, et~al.
\newblock The rise and potential of large language model based agents: A survey.
\newblock {\em arXiv preprint arXiv:2309.07864}, 2023.

\bibitem{brohan2023can}
Anthony Brohan, Yevgen Chebotar, Chelsea Finn, Karol Hausman, Alexander Herzog, Daniel Ho, Julian Ibarz, Alex Irpan, Eric Jang, Ryan Julian, et~al.
\newblock Do as i can, not as i say: Grounding language in robotic affordances.
\newblock In {\em Conference on Robot Learning}, pages 287--318. PMLR, 2023.

\bibitem{wei2022chain}
Jason Wei, Xuezhi Wang, Dale Schuurmans, Maarten Bosma, Fei Xia, Ed~Chi, Quoc~V Le, Denny Zhou, et~al.
\newblock Chain-of-thought prompting elicits reasoning in large language models.
\newblock {\em Advances in Neural Information Processing Systems}, 35:24824--24837, 2022.

\bibitem{koh2024visualwebarena}
Jing~Yu Koh, Robert Lo, Lawrence Jang, Vikram Duvvur, Ming~Chong Lim, Po-Yu Huang, Graham Neubig, Shuyan Zhou, Ruslan Salakhutdinov, and Daniel Fried.
\newblock Visualwebarena: Evaluating multimodal agents on realistic visual web tasks.
\newblock {\em arXiv preprint arXiv:2401.13649}, 2024.

\bibitem{deng2023mind2web}
Xiang Deng, Yu~Gu, Boyuan Zheng, Shijie Chen, Samuel Stevens, Boshi Wang, Huan Sun, and Yu~Su.
\newblock Mind2web: Towards a generalist agent for the web.
\newblock {\em arXiv preprint arXiv:2306.06070}, 2023.

\bibitem{gur2023html}
Izzeddin Gur, Ofir Nachum, Yingjie Miao, Mustafa Safdari, Austin Huang, Aakanksha Chowdhery, Sharan Narang, Noah Fiedel, and Aleksandra Faust.
\newblock Understanding {HTML} with large language models.
\newblock In Houda Bouamor, Juan Pino, and Kalika Bali, editors, {\em Findings of the Association for Computational Linguistics: EMNLP 2023}, pages 2803--2821, Singapore, December 2023. Association for Computational Linguistics.

\bibitem{zhou2023webarena}
Shuyan Zhou, Frank~F Xu, Hao Zhu, Xuhui Zhou, Robert Lo, Abishek Sridhar, Xianyi Cheng, Yonatan Bisk, Daniel Fried, Uri Alon, et~al.
\newblock Webarena: A realistic web environment for building autonomous agents.
\newblock {\em arXiv preprint arXiv:2307.13854}, 2023.

\bibitem{lu-etal-2022-fantastically}
Yao Lu, Max Bartolo, Alastair Moore, Sebastian Riedel, and Pontus Stenetorp.
\newblock Fantastically ordered prompts and where to find them: Overcoming few-shot prompt order sensitivity.
\newblock In {\em ACL}, 2022.

\bibitem{dong2022survey}
Qingxiu Dong, Lei Li, Damai Dai, Ce~Zheng, Zhiyong Wu, Baobao Chang, Xu~Sun, Jingjing Xu, and Zhifang Sui.
\newblock A survey for in-context learning.
\newblock {\em arXiv preprint arXiv:2301.00234}, 2022.

\bibitem{min-etal-2022-rethinking}
Sewon Min, Xinxi Lyu, Ari Holtzman, Mikel Artetxe, Mike Lewis, Hannaneh Hajishirzi, and Luke Zettlemoyer.
\newblock Rethinking the role of demonstrations: What makes in-context learning work?
\newblock In {\em EMNLP}, 2022.

\bibitem{kaddour2023challenges}
Jean Kaddour, Joshua Harris, Maximilian Mozes, Herbie Bradley, Roberta Raileanu, and Robert McHardy.
\newblock Challenges and applications of large language models, 2023.

\bibitem{zheng2023seeact}
Boyuan Zheng, Boyu Gou, Jihyung Kil, Huan Sun, and Yu~Su.
\newblock Gpt-4v(ision) is a generalist web agent, if grounded.
\newblock {\em arXiv preprint arXiv:2401.01614}, 2024.

\bibitem{zeng2023agenttuning}
Aohan Zeng, Mingdao Liu, Rui Lu, Bowen Wang, Xiao Liu, Yuxiao Dong, and Jie Tang.
\newblock Agenttuning: Enabling generalized agent abilities for llms.
\newblock {\em arXiv preprint arXiv:2310.12823}, 2023.

\bibitem{huang2022language}
Wenlong Huang, Pieter Abbeel, Deepak Pathak, and Igor Mordatch.
\newblock Language models as zero-shot planners: Extracting actionable knowledge for embodied agents.
\newblock In {\em International Conference on Machine Learning}, pages 9118--9147. PMLR, 2022.

\bibitem{logeswaran-etal-2022-shot}
Lajanugen Logeswaran, Yao Fu, Moontae Lee, and Honglak Lee.
\newblock Few-shot subgoal planning with language models.
\newblock In {\em NAACL: HLT}, 2022.

\bibitem{gao2023pal}
Luyu Gao, Aman Madaan, Shuyan Zhou, Uri Alon, Pengfei Liu, Yiming Yang, Jamie Callan, and Graham Neubig.
\newblock Pal: Program-aided language models.
\newblock In {\em International Conference on Machine Learning}, pages 10764--10799. PMLR, 2023.

\bibitem{qin2023toolllm}
Yujia Qin, Shihao Liang, Yining Ye, Kunlun Zhu, Lan Yan, Yaxi Lu, Yankai Lin, Xin Cong, Xiangru Tang, Bill Qian, et~al.
\newblock Toolllm: Facilitating large language models to master 16000+ real-world apis.
\newblock {\em arXiv preprint arXiv:2307.16789}, 2023.

\bibitem{patil2023gorilla}
Shishir~G. Patil, Tianjun Zhang, Xin Wang, and Joseph~E. Gonzalez.
\newblock Gorilla: Large language model connected with massive apis.
\newblock {\em arXiv preprint arXiv:2305.15334}, 2023.

\bibitem{parisi2022talm}
Aaron Parisi, Yao Zhao, and Noah Fiedel.
\newblock Talm: Tool augmented language models.
\newblock {\em arXiv preprint arXiv:2205.12255}, 2022.

\bibitem{schick2023toolformer}
Timo Schick, Jane Dwivedi-Yu, Roberto Dess{\`\i}, Roberta Raileanu, Maria Lomeli, Luke Zettlemoyer, Nicola Cancedda, and Thomas Scialom.
\newblock Toolformer: Language models can teach themselves to use tools.
\newblock {\em arXiv preprint arXiv:2302.04761}, 2023.

\bibitem{sun2023adaplanner}
Haotian Sun, Yuchen Zhuang, Lingkai Kong, Bo~Dai, and Chao Zhang.
\newblock Adaplanner: Adaptive planning from feedback with language models.
\newblock {\em arXiv preprint arXiv:2305.16653}, 2023.

\bibitem{yao2022react}
Shunyu Yao, Jeffrey Zhao, Dian Yu, Nan Du, Izhak Shafran, Karthik Narasimhan, and Yuan Cao.
\newblock React: Synergizing reasoning and acting in language models.
\newblock {\em arXiv preprint arXiv:2210.03629}, 2022.

\bibitem{shridhar20alfworld}
Mohit Shridhar, Xingdi Yuan, Marc-Alexandre C\^ot\'e, Yonatan Bisk, Adam Trischler, and Matthew Hausknecht.
\newblock {ALFWorld: Aligning Text and Embodied Environments for Interactive Learning}.
\newblock In {\em ICLR}, 2021.

\bibitem{madaan2023self}
Aman Madaan, Niket Tandon, Prakhar Gupta, Skyler Hallinan, Luyu Gao, Sarah Wiegreffe, Uri Alon, Nouha Dziri, Shrimai Prabhumoye, Yiming Yang, et~al.
\newblock Self-refine: Iterative refinement with self-feedback.
\newblock {\em arXiv preprint arXiv:2303.17651}, 2023.

\bibitem{kim2023language}
Geunwoo Kim, Pierre Baldi, and Stephen McAleer.
\newblock Language models can solve computer tasks.
\newblock {\em arXiv preprint arXiv:2303.17491}, 2023.

\bibitem{shinn2023reflexion}
Noah Shinn, Federico Cassano, Ashwin Gopinath, Karthik~R Narasimhan, and Shunyu Yao.
\newblock Reflexion: Language agents with verbal reinforcement learning.
\newblock In {\em Thirty-seventh Conference on Neural Information Processing Systems}, 2023.

\bibitem{branavan2012learning}
SRK Branavan, David Silver, and Regina Barzilay.
\newblock Learning to win by reading manuals in a monte-carlo framework.
\newblock {\em Journal of Artificial Intelligence Research}, 43:661--704, 2012.

\bibitem{hanjie2021grounding}
Austin~W Hanjie, Victor~Y Zhong, and Karthik Narasimhan.
\newblock Grounding language to entities and dynamics for generalization in reinforcement learning.
\newblock In {\em International Conference on Machine Learning}, pages 4051--4062. PMLR, 2021.

\bibitem{zhong2020rtfm}
Victor Zhong, Tim Rockt{\"a}schel, and Edward Grefenstette.
\newblock Rtfm: Generalising to new environment dynamics via reading.
\newblock In {\em ICLR}, pages 1--17. ICLR, 2020.

\bibitem{zhao2023expel}
Andrew Zhao, Daniel Huang, Quentin Xu, Matthieu Lin, Yong-Jin Liu, and Gao Huang.
\newblock Expel: Llm agents are experiential learners.
\newblock {\em arXiv preprint arXiv:2308.10144}, 2023.

\bibitem{sutton1998rlbook}
Richard~S. Sutton and Andrew~G. Barto.
\newblock {\em Reinforcement Learning: An Introduction}.
\newblock The MIT Press, second edition, 2018.

\bibitem{yao2022webshop}
Shunyu Yao, Howard Chen, John Yang, and Karthik Narasimhan.
\newblock Webshop: Towards scalable real-world web interaction with grounded language agents.
\newblock {\em Advances in Neural Information Processing Systems}, 35:20744--20757, 2022.

\bibitem{shridhar20alfred}
Mohit Shridhar, Jesse Thomason, Daniel Gordon, Yonatan Bisk, Winson Han, Roozbeh Mottaghi, Luke Zettlemoyer, and Dieter Fox.
\newblock {ALFRED: A Benchmark for Interpreting Grounded Instructions for Everyday Tasks}.
\newblock In {\em The IEEE Conference on Computer Vision and Pattern Recognition (CVPR)}, 2020.

\bibitem{gtravel}
{Google Flights}.
\newblock \url{https://www.google.com/travel/flights}.
\newblock Accessed: 2024-05-21.

\bibitem{yang2023setofmark}
Jianwei Yang, Hao Zhang, Feng Li, Xueyan Zou, Chunyuan Li, and Jianfeng Gao.
\newblock Set-of-mark prompting unleashes extraordinary visual grounding in gpt-4v.
\newblock {\em arXiv preprint arXiv:2310.11441}, 2023.

\end{thebibliography}

%%%%%%%%%%%%%%%%%%%%%%%%%%%%%%%%%%%%%%%%%%%%%%%%%%%%%%%%%%%%
% Appendix
\appendix
\clearpage
\appendix
\section{Limitation and Broader Impacts}\label{sec:limitation-broader-impact}
\textbf{Limitation.} The performance of \textsc{\Ours} depends on the diversity of offline experiences. As such, one important direction for improvement is to automatically collect diverse offline experiences through continual learning, where we iteratively generate guidelines and use them to gather more trajectories with high rewards. Another avenue is the need for quantifying the quality of generated contexts and guidelines. Currently, apart from applying context-aware guidelines to ReAct and measuring test time performance, there lacks a standardized method for quantifying the quality of generated contexts and selected guidelines. Introducing a quantifiable metric to approximate the quality could pave the way for new optimization approaches such as reinforcement learning.

\textbf{Broader impact.} This paper introduces research aimed at enhancing the decision-making capabilities of LLMs. In terms of societal impact, while we develop a generic LLM-based autonomous agent, having biased offline datasets may lead to making decisions with suboptimal outcomes. Additionally, autonomous agents may be misused for malicious applications. To mitigate these risks, potential solutions would include diversifying datasets, implementing ethical oversight, ensuring transparency and accountability, engaging with stakeholders for a broader perspective, and incorporating security measures to prevent misuse. We believe that the research community, including ourselves, should responsibly advance LLM-based agent research, prioritizing societal well-being and ethical considerations.

\section{Evaluation Details}\label{sec:appendix-evaluation-details}
\subsection{ALFWorld \cite{shridhar20alfworld}}\label{sec:appendix-env-details}
\subsubsection{Environment details}
Each task in ALFWorld starts with a description of the specific environment and the goal to achieve. At each timestep, an agent can choose one of the following actions to interact with the objects and receptacles in the environment: 
\begin{itemize}
\itemsep0em
\item  go to [recep]
\item  take [object] from [recep] 
\item  put [object] in/on [recep]
\item  open/close/use [recep]
\item  clean/heat/cool [object] with [recep]
\end{itemize}
Alternatively, the agent can generate a think action for planning and reflection, which helps with decision-making but does not change the environment itself.
After one action is performed, the environment returns an observation that describes view changes.

Following ReAct, we concatenate a list of (observation, action) pairs to show the entire trajectory up to the current timestep for LLM agents to generate the next action. We experiment on 134 unseen test tasks with 6 categories of \texttt{pick\_and\_place}, \texttt{pick\_clean\_then\_place}, \texttt{pick\_heat\_then\_place}, \texttt{pick\_cool\_then\_place}, \texttt{look\_at\_obj}, and \texttt{pick\_two\_obj}.
For each task, the agent is allowed to take a maximum of 50 actions.

\subsubsection{Baseline and Models}
For ALFWorld tasks, we follow the same setting as ReAct by providing 2 in-context examples for each of the 6 task categories. The original results of ReAct in their paper are produced based on Text-Davinci-002. However, this GPT version is no longer available, so we apply gpt-3.5-turbo-instruct instead to generate actions. For ExpeL, we directly take the guidelines from their appendix and append them to the ReAct agent at test time. For Reflexion, we generate reflective feedback based on the test rewards if the tasks fail. This reflective feedback is appended to the action generation prompt, immediately following the context-aware guidelines for AutoGuide + Reflexion. When there are multiple past episodes, we concatenate the reflection feedback from each episode. This process continues until the maximum number of episodes (or trials) is reached.

\subsubsection{Implementation details of \textsc{\Ours}}
We run the first 100 training tasks of ALFWorld to collect $(\bm{\tau}_{+},\bm{\tau}_{-})$ pairs with ReAct+Reflexion and extract context-dependant guidelines on the collected data. For context identification, we provide 2-shot demonstrations for each of the 6 task categories.
The corresponding prompt templates can be found in \cref{sec:appendix-prompt-template}.
All parameter details are shown in \cref{tab:exp_parameter_alfworld}.

\begin{table}[!ht]
\centering
\begin{tabular}{c|c}
\toprule
\makecell{Parameter name} & \makecell{Value} \\ 
\midrule
Allowed Episode Length & 50 \\ 
% \hline
n-shots & 2 \\ 
% \# training tasks & 100 \\ 
Agent Model & gpt-3.5-turbo-instruct \\
Context Identification Model & gpt-3.5-turbo-instruct \\
Guideline Selection Model & gpt-3.5-turbo-instruct \\
Guideline Extraction Model & gpt-4-1106-preview \\
Reflexion Model & gpt-4-1106-preview \\
top-k guideline selection  & 2 \\
\bottomrule
\end{tabular}
\vspace{5pt}
\caption{Experiment hyperparameters on ALFWorld.  The maximum
allowed episode length and
n-shots follow the same setup in ReAct.}
\label{tab:exp_parameter_alfworld}
\end{table}

\subsection{WebShop \cite{yao2022webshop}}
\subsubsection{Environment details}
WebShop provides an e-commerce environment, where the objective is to find and buy the product that matches the task-specific Instruction.
The agent can select one of the following actions to perform: 
\begin{itemize}
\itemsep0em
\item search[query]
\item click[button]
\end{itemize}
Following ReAct, the agent can generate think actions to do planning or reflection. After buying a product, the environment returns a reward showing how well the bought product matches the target one in type, price, buying options, and attributes. The reward is calculated by:
$$r\!=\!r_{type} \cdot \frac{|U_{att} \cap Y_{att}|\!+\!|U_{opt} \cap Y_{opt}|\!+\!\mathbbm{1}[y_{price} \leq u_{price}]}{|U_{att}|\!+\!|U_{opt}|\!+\!1}$$
where y is the bought product and u is the target product. Same as ALFWorld, for WebShop, the agent takes ($\mathrm{obs}_t$, $\mathrm{act}_t$) pairs for every previous timestep $t$ as input to generate the next action. 

\subsubsection{Baseline and Models}
Following ReAct, experiments are done in a one-shot setting. We apply gpt-3.5-turbo-0613 to generate actions, but when the token number exceeds the token limit (for example, for the n-shot ReAct experiments in \cref{table:main-result}), we use the 16k version of gpt-3.5-turbo-0613 instead. 
For ExpeL, we could not find how many training tasks the framework used for training. Therefore, we directly apply the guidelines from the appendix of their paper at test time. We only consider ExpeL with guidelines, not ExpeL with in-context example selection in our experiments for a fair comparison. The in-context example selection method is orthogonal to our work and can be easily combined with our method.
For Reflexion, as shown in their paper, their 2-shot Reflexion prompt does not work well on WebShop. Therefore, we re-write the prompt and apply gpt-4-1106-preview to generate episode-level reflections for all Reflexion experiments. Same as the experiments in ALFWorld, the generated reflections are appended to the action generation prompt, immediately following the context-aware guidelines for AutoGuide + Reflexion. In cases where multiple past episodes are available, the reflection feedback from each episode is concatenated. This process is repeated until the maximum number of trials is reached.
Following Reflexion and ExpeL, the evaluation is done on 100 test tasks. The maximum number of allowed actions for each task is 15. At the same time, each search action shows the top 3 products for the search query. Please refer to \Cref{tab:exp_parameter_webshop} for more details about the experiments.

\subsubsection{Implementation details of \textsc{\Ours}}
We randomly sample 50 training tasks from the training set of WebShop, on which we run ReAct+Reflexion to collect pairs and generate guidelines. 
The context identification prompt is one-shot, which is shown in \Cref{sec:appendix-prompt-template}.
At test time, we ask gpt-3.5-turbo-0613 to select each state's most relevant top 2 guidelines. 

\begin{table}[!ht]
\centering
\begin{tabular}{c|c}
\toprule
\makecell{Parameter name} & \makecell{Value} \\ 
\midrule
Allowed Episode Length & 15 \\
\# of Search Results & 3 \\
% \hline
n-shots & 1 \\ 
Agent Model & gpt-3.5-turbo-0613 \\
Context Identification Model & gpt-3.5-turbo-0613 \\
Guideline Selection Model & gpt-3.5-turbo-0613 \\
Guideline Extraction Model & gpt-4-1106-preview \\
Reflexion Model & gpt-4-1106-preview \\
top-k guideline selection & 2 \\
\bottomrule
\end{tabular}
\vspace{5pt}
\caption{Experiment hyperparameters on WebShop. The maximum allowed episode length, the number of search results per page, and n-shots follow the same setup in ReAct.}
\label{tab:exp_parameter_webshop}
\end{table}

\subsection{WebArena \cite{zhou2023webarena}}
\subsubsection{Environment details}
WebArena provides web-based benchmark environments that closely follow the data and functionality of real-world websites. 
Unlike other benchmarks like WebShop that provide clean text of website information as observation, WebArena's webpage content is represented as an accessibility tree, which is a subset of the DOM tree with useful elements of a webpage. 
For our expeirments, we focus on WebArena Reddit, which simulates the real Reddit websites with users, forums, and posts with abundant text information.

For each task in WebArena, the agent is expected to achieve a task-specific intent. At each timestep, WebArena provides 
a list of opened tabs, 
the accessibility tree of the focused webpage, 
and the URL of the current page
as observation. 
For WebArena, each observation is long. Therefore, following the baseline in WebArena, at each timestep, we only provide the observation of the current timestep to the agent.
We additionally provide up to 5 past actions for the agent to understand what it did in the past. 
The allowed actions in WebArena include the following: 
\begin{itemize}
\itemsep0em
\item goto [url]
\item click [element\_id]
\item type [element\_id] [text] [1 for enter or 0 for not enter]
\item press [key\_combination]
\item scroll [up or down]
\item go\_back
\end{itemize}
The maximum allowed number of actions for a single task is 20. Note that WebArena does not provide training tasks, but the work provides 19 demonstrations for Reddit, each of a different category. Therefore, we set these 19 tasks as the training tasks and then test on the rest 87 tasks. 

\subsubsection{Baseline and Models}
We directly run the two-shot ReAct-style baseline in the official codebase of WebArena using gpt4-preview-1106. 
For ExpeL, the original paper does not include experiments on WebArena, therefore we try our best to implement our own version and run on the same training tasks as our method.

\subsubsection{Implementation details of \textsc{\Ours}}
% Among the 19 human demonstrations provided in WebArena, 17 of them successfully complete the tasks. 
We directly run ReAct on the tasks with $\bm{\tau}_{+}$ to collect $\bm{\tau}_{-}$ actions and generate guidelines correspondingly. 
We provide a 5-shot prompt for the context identification module, which is shown in \Cref{fig:guideline-prompt-state-summarization-webarena}. At test time, the top 2 guidelines at each timestep are selected to guide action generation.

% We take advantage of the URL and webpage titles, which are a part of the observation for this benchmark, to help with context identification.

\begin{table}[!ht]
\centering
\begin{tabular}{c|c}
\toprule
\makecell{Parameter name} & \makecell{Value} \\ 
\midrule
Allowed Episode Length & 20 \\ 
% \hline
n-shots & 2 \\
Agent Model & gpt-4-1106-preview \\
Context Identification Model & gpt-4-1106-preview \\
Guideline Selection Model & gpt-4-1106-preview \\
Guideline Extraction Model & gpt-4-1106-preview \\
top-k guideline selection & 2 \\
\bottomrule
\end{tabular}
\vspace{5pt}
\caption{Experiment hyperparameters on WebArena. The number of shots follows the same setup in ReAct.}
\label{tab:exp_parameter_webarena}
\end{table}

\subsection{Real Websites}
\label{sec:appendix_real_website_details}
\subsubsection{Environment details}
We design a set of real-world website navigation tasks from 3 domains: Software Development (GitHub), Travel (Google Flights), and Education (Coursera), which have 30, 20, and 20 test tasks accordingly. Here are some example tasks:
\begin{itemize}
    \item GitHub
    \begin{itemize}
        \item Navigate to the repository for the Python library Seaborn with the most stars and show me all the open issues labeled with bug.
        \item Go to the GitHub org for Spotify and open the pinned project with the most stars for me.
    \end{itemize}
    \item Google Flights
    \begin{itemize}
        \item Find me the one-way flight from Hong Kong to Paris departing on Oct 15th 2024 with the least emmisions.
        \item Show me the booking options of the one-way flight departing from Auckland on September 11, 2024, and arriving in Rome with the earliest departure time on that day.
    \end{itemize}
    \item Coursera
    \begin{itemize}
        \item Show me a Cybersecurity course that can finish within 1 month and show me all the reviews for the selected course.
        \item Find me a Coursera guided project that covers Unity and show me its main page.
    \end{itemize}
\end{itemize}
We follow the action space design of Visual WebArena~\cite{koh2024visualwebarena}, which has the following action types available: 
\begin{itemize}
\itemsep0em
\item goto [url]
\item click [element\_id]
\item hover [element\_id]
\item type [element\_id] [text] [1 for enter or 0 for not enter]
\item press [key\_combination]
\item scroll [up or down]
\item tab\_focus [tab\_index]
\item close\_tab
\item go\_back
\item go\_forward
\end{itemize}
As the real websites are constantly and dynamically changing, we evaluate the completed task with human experts. 
\subsubsection{Baseline and Models}
We directly run the two-shot SoM algorithm in the official codebase of Visual WebArena. The only modification we made from the original codebase is the bounding box detection algorithm, in which we further filter invisible bounding boxes from the list and add the list elements as interactable elements to consider. 
\subsubsection{Implementation details of \textsc{\Ours}}
We provide a total of 6 training tasks (3 for GitHub, 2 for Google Travel, and 1 for Coursera), on which we collect human demonstration as $\bm{\tau}_{+}$'s and run SoM to collect $\bm{\tau}_{-}$'s. From the pairs with multi-modal observations we generate text-based guidelines to guide action selection. Both context identification and guideline extraction are done by gpt-4-vision-preview, and we provide a 3-shot prompt for context identification. All parameter details are shown in \Cref{tab:exp_parameter_multimodal}. 
\begin{table}[!ht]
\centering
\begin{tabular}{c|c}
\toprule
\makecell{Parameter name} & \makecell{Value} \\ 
\midrule
Allowed Episode Length & 15 \\
n-shots & 2 \\ 
Agent Model & gpt-4-vision-preview \\
Context Identification Model & gpt-4-vision-preview \\
Guideline Selection Model & gpt-4-vision-preview \\
Guideline Extraction Model & gpt-4-vision-preview \\
top-k guideline selection & 2 \\
\bottomrule
\end{tabular}
\vspace{5pt}
\caption{Experiment hyperparameters on multi-modal real-world website tasks.}
\label{tab:exp_parameter_multimodal}
\end{table}

\section{Prompt Templates}\label{sec:appendix-prompt-template}
\subsection{Context Identification}\label{sec:appendix-prompt-template-context-summarization}

%%%%%%%%%%%%%%%%%%%%%%%%%%%%%%%%%%%%%%%%%%%%%%%%%%%%%%%%%
\begin{figure*}[t!]
    \centering
    \includegraphics[width=\textwidth]{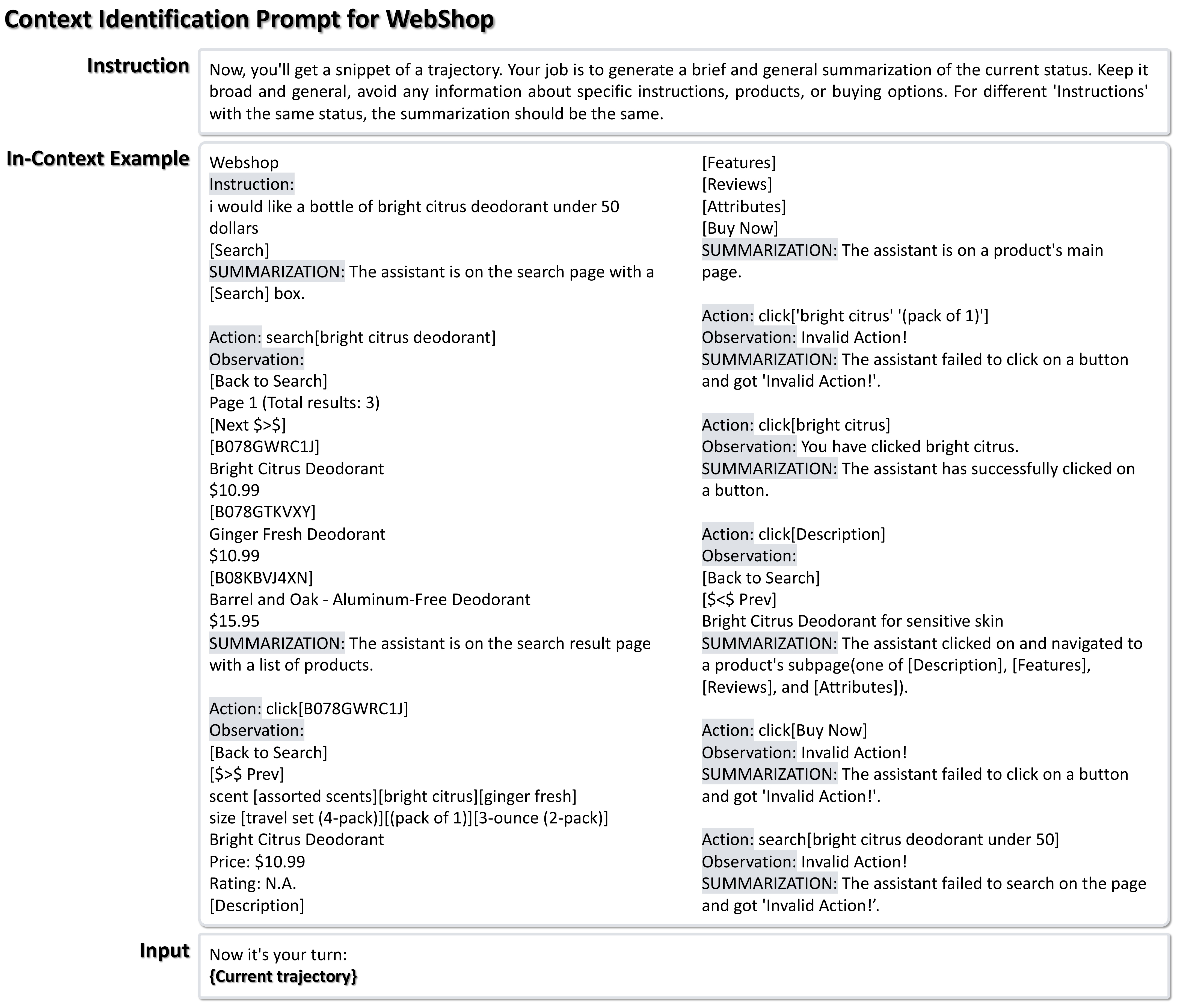}
    \caption{Our prompt template for context identification (\Cref{eqn:context-summarization-module}) in the WebShop domain.}
    \label{fig:guideline-prompt-state-summarization-webshop}
\end{figure*}
%%%%%%%%%%%%%%%%%%%%%%%%%%%%%%%%%%%%%%%%%%%%%%%%%%%%%%%%%

%%%%%%%%%%%%%%%%%%%%%%%%%%%%%%%%%%%%%%%%%%%%%%%%%%%%%%%%%
\begin{figure*}[t!]
    \centering
    \includegraphics[width=\textwidth]{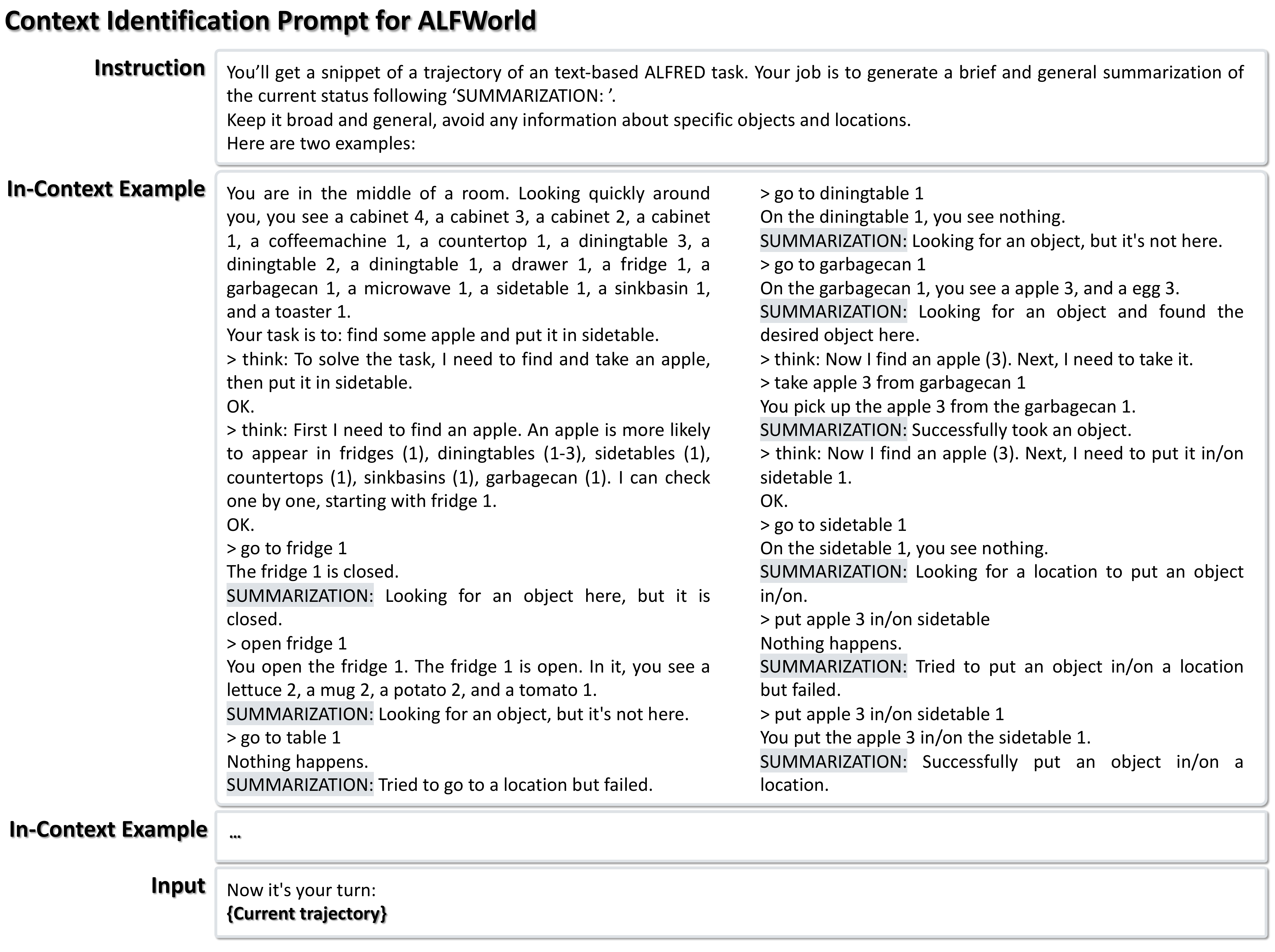}
    \caption{Our prompt template for context identification (\Cref{eqn:context-summarization-module}) in the ALFWorld domain.}
    \label{fig:guideline-prompt-state-summarization-alfworld}
\end{figure*}
%%%%%%%%%%%%%%%%%%%%%%%%%%%%%%%%%%%%%%%%%%%%%%%%%%%%%%%%%

%%%%%%%%%%%%%%%%%%%%%%%%%%%%%%%%%%%%%%%%%%%%%%%%%%%%%%%%%
\begin{figure*}[t!]
    \centering
    \includegraphics[width=\textwidth]{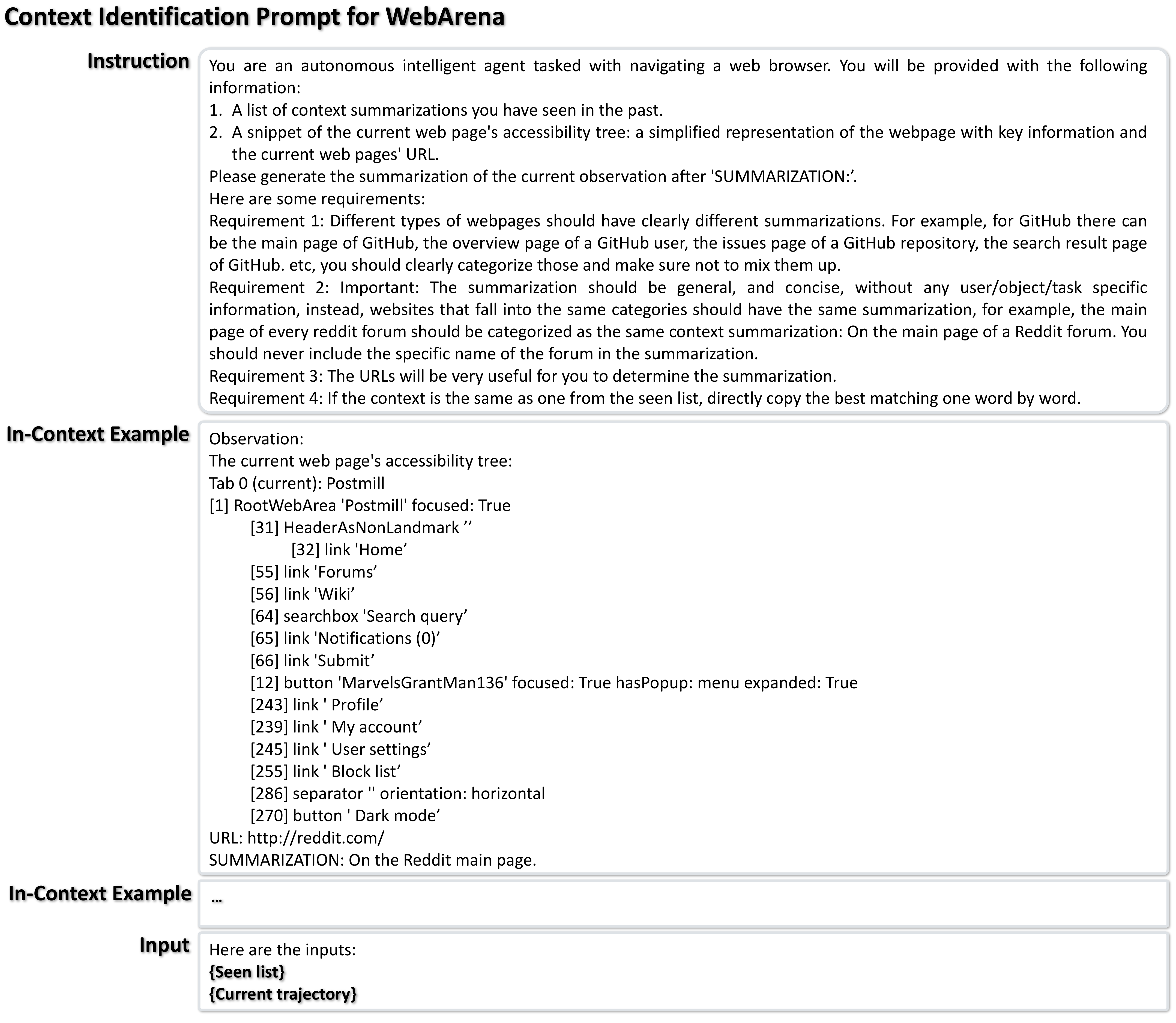}
    \caption{Our prompt template for context identification (\Cref{eqn:context-summarization-module}) in the WebArena domain.}
    \label{fig:guideline-prompt-state-summarization-webarena}
    \vspace{1cm}
\end{figure*}
%%%%%%%%%%%%%%%%%%%%%%%%%%%%%%%%%%%%%%%%%%%%%%%%%%%%%%%%%

We present our prompt templates for the context identification module $\mathcal{M}_{\mathrm{context}}$ (\Cref{eqn:context-summarization-module}) for WebShop, ALFWorld, and WebArena in \Cref{fig:guideline-prompt-state-summarization-webshop,fig:guideline-prompt-state-summarization-alfworld,fig:guideline-prompt-state-summarization-webarena}, respectively.
In ALFWorld, there exist six categories of tasks, and we use context identification prompting with 2-shot examples for each task, following the practice by \cite{yao2022react}. \Cref{fig:guideline-prompt-state-summarization-alfworld} shows one example for the \texttt{pick\_and\_place} tasks.

\subsection{Guideline Extraction}\label{sec:appendix-prompt-template-guideline-extraction}

\begin{figure}[ht]
    \centering
    \includegraphics[width=\textwidth]{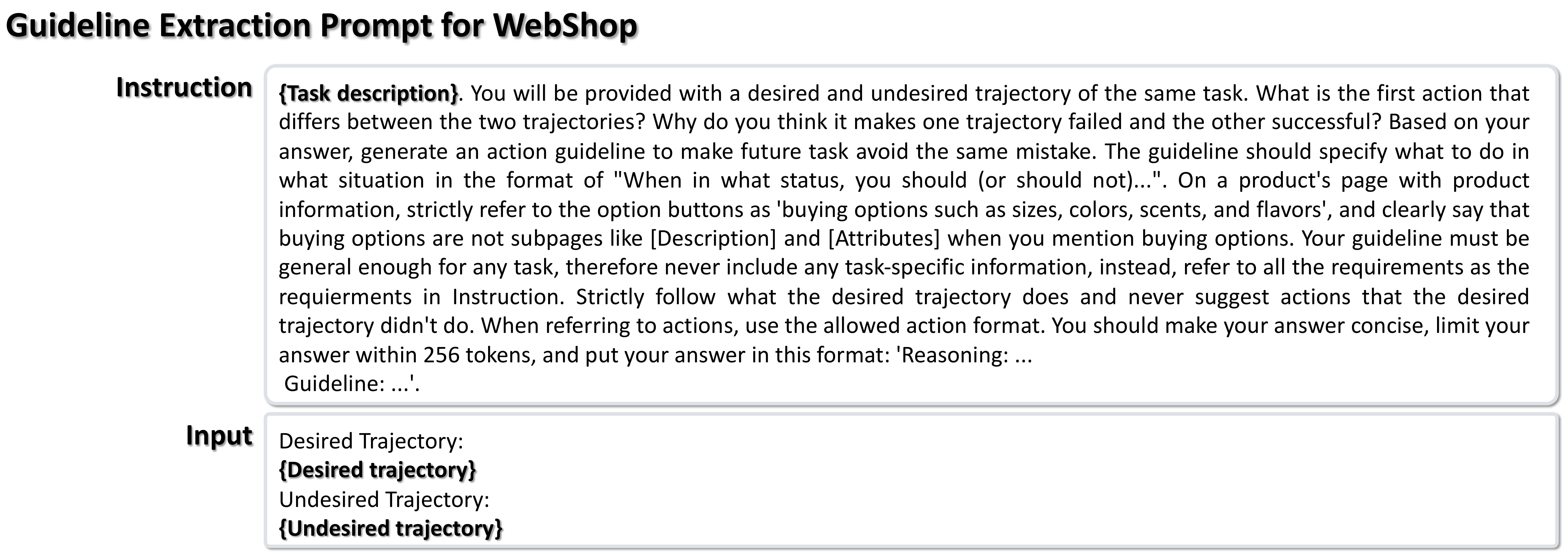}
    \caption{Our prompt template for guideline extraction (\Cref{eqn:action-extraction-module}) in the WebShop domain.}
    \label{fig:guideline-prompt-guide-extraction-webshop}
\end{figure}

\begin{figure*}[t!]
    \centering
    \includegraphics[width=\textwidth]{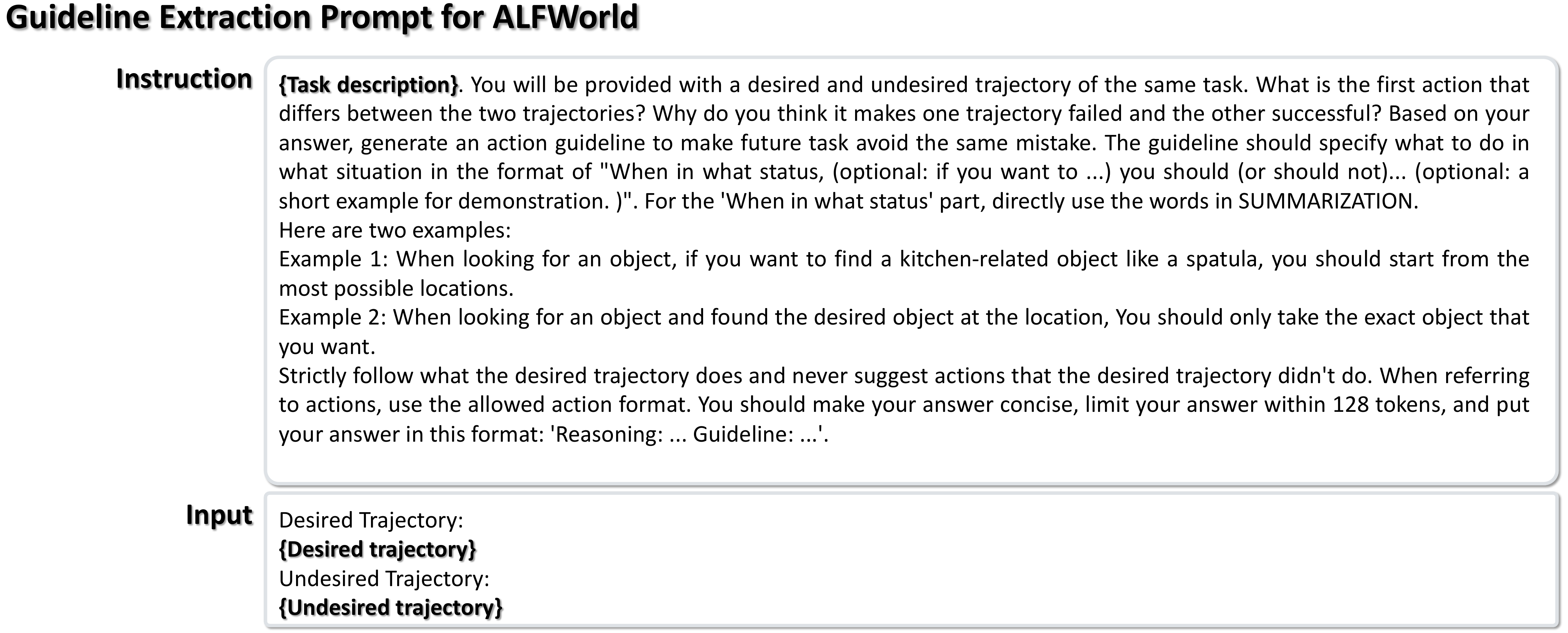}
    \caption{Our prompt template for guideline extraction (\Cref{eqn:action-extraction-module}) in the ALFWorld domain.}
    \label{fig:guideline-prompt-guide-extraction-alfworld}
\end{figure*}

\begin{figure*}[t!]
    \centering
    \includegraphics[width=\textwidth]{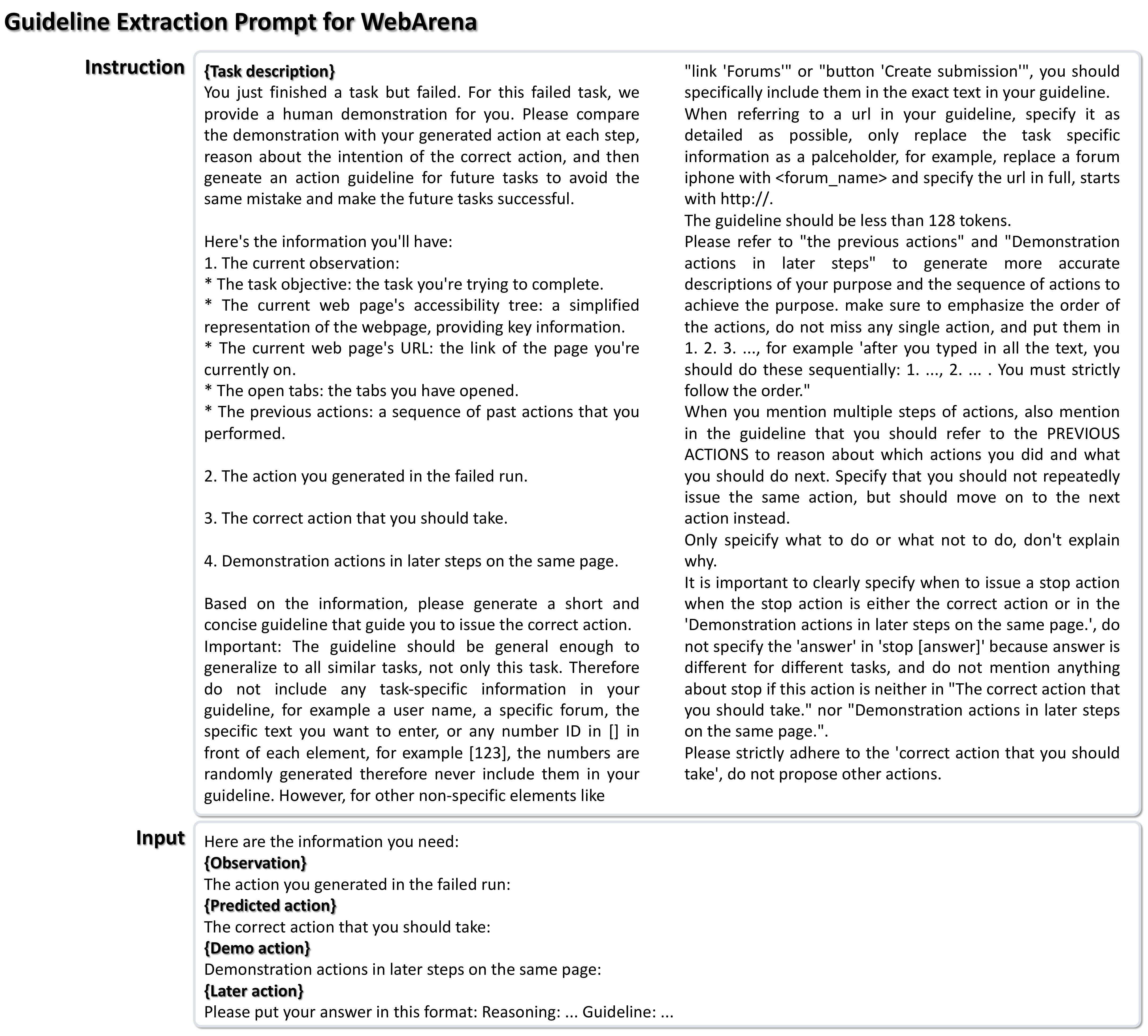}
    \caption{Our prompt template for guideline extraction (\Cref{eqn:action-extraction-module}) in the WebArena domain.}
    \label{fig:guideline-prompt-guide-extraction-webarena}
\end{figure*}

\Cref{fig:guideline-prompt-guide-extraction-webshop,fig:guideline-prompt-guide-extraction-alfworld,fig:guideline-prompt-guide-extraction-webarena} detail our prompt templates $\mathcal{M}_{\mathrm{guideline}}$ for extracting guidelines (\Cref{eqn:action-extraction-module}) in the WebShop, ALFWorld, and WebArena domains.

%%%%%%%%%%%%%%%%%%%%%%%%%%%%%%%%%%%%%%%%%%%%%%%%%%%%%%%%%
%%%%%%%%%%%%%%%%%%%%%%%%%%%%%%%%%%%%%%%%%%%%%%%%%%%%%%%%%

%%%%%%%%%%%%%%%%%%%%%%%%%%%%%%%%%%%%%%%%%%%%%%%%%%%%%%%%%
%%%%%%%%%%%%%%%%%%%%%%%%%%%%%%%%%%%%%%%%%%%%%%%%%%%%%%%%%

%%%%%%%%%%%%%%%%%%%%%%%%%%%%%%%%%%%%%%%%%%%%%%%%%%%%%%%%%
%%%%%%%%%%%%%%%%%%%%%%%%%%%%%%%%%%%%%%%%%%%%%%%%%%%%%%%%%

\subsection{Context Matching}\label{sec:appendix-prompt-template-context-matching}

%%%%%%%%%%%%%%%%%%%%%%%%%%%%%%%%%%%%%%%%%%%%%%%%%%%%%%%%%
\begin{figure*}[t!]
    \centering
    \includegraphics[width=\textwidth]{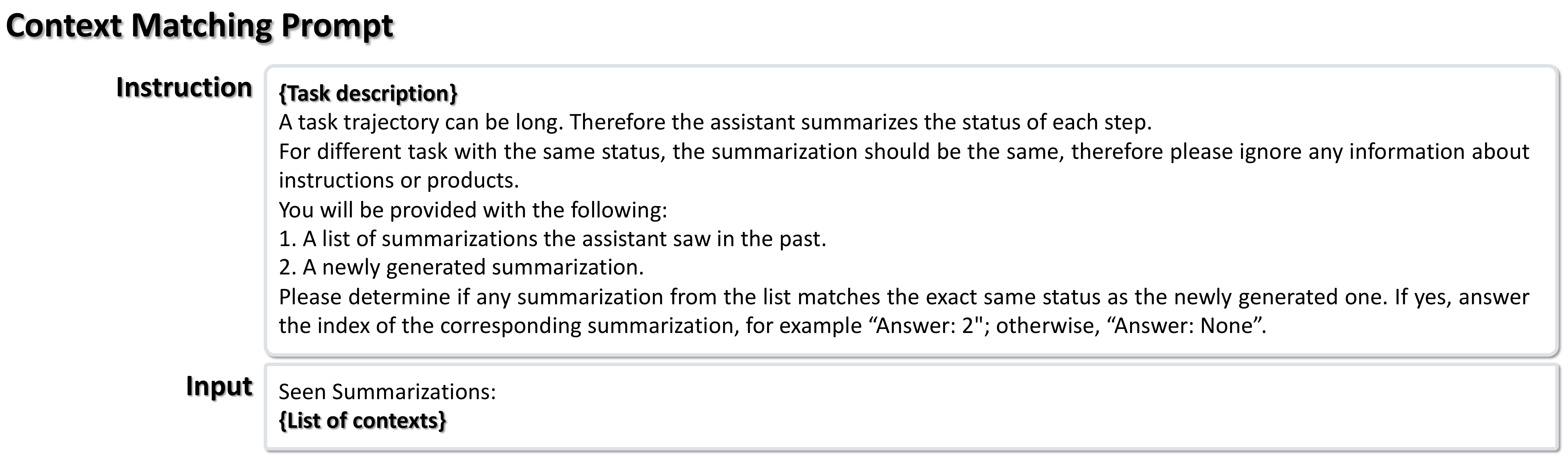}
    \caption{Our prompt template for context matching (\Cref{sec:extraction-context-aware-guidelines,sec:applying-context-aware-guidelines}) in all the WebShop, ALFWorld, and WebArena domains.}
    \label{fig:guideline-prompt-state-categorization}
\end{figure*}
%%%%%%%%%%%%%%%%%%%%%%%%%%%%%%%%%%%%%%%%%%%%%%%%%%%%%%%%%

\Cref{fig:guideline-prompt-state-categorization} shows our prompt template for matching the generated context with one of the existing contexts if there is any similar context, for the construction of the set of context-aware guidelines (\Cref{sec:extraction-context-aware-guidelines}) and the retrieval of relevant guidelines during testing (\Cref{sec:applying-context-aware-guidelines}) in all the three domains: WebShop, ALFWorld, and WebArena.

\subsection{Guideline Selection}\label{sec:appendix-prompt-template-guideline-retrieval}

%%%%%%%%%%%%%%%%%%%%%%%%%%%%%%%%%%%%%%%%%%%%%%%%%%%%%%%%%
\begin{figure*}[t!]
    \centering
    \includegraphics[width=\textwidth]{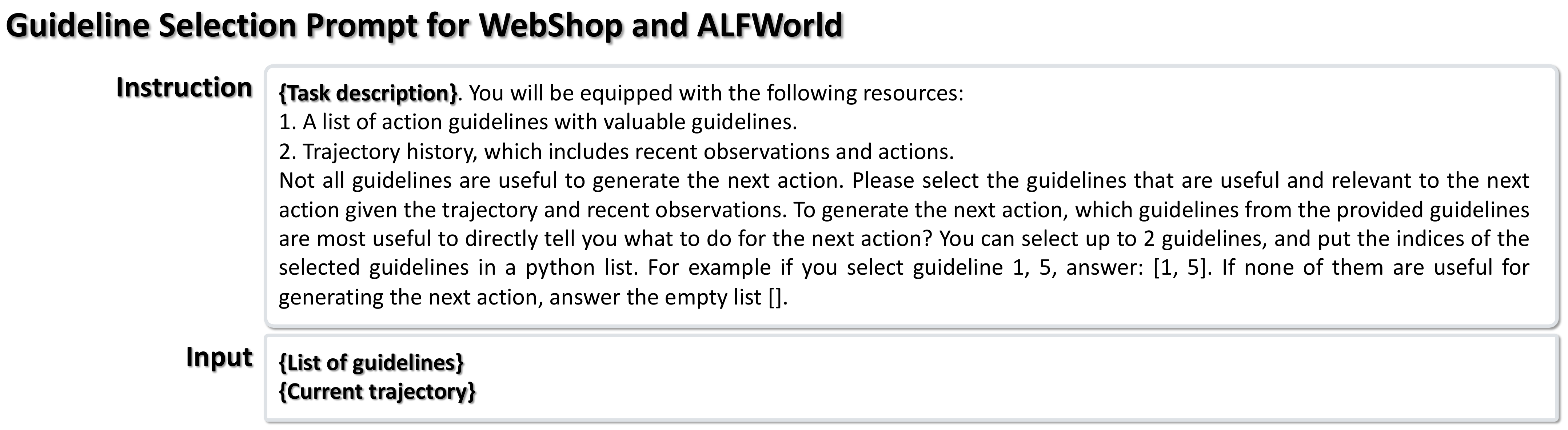}
    \caption{Our prompt template for selecting the most relevant context-aware guidelines during the test time (\Cref{eqn:guideline-selection-module} from \Cref{sec:applying-context-aware-guidelines}) in the WebShop and ALFWorld domains.}
    \label{fig:guideline-prompt-guide-retrieval-webshop-alfworld}
\end{figure*}
%%%%%%%%%%%%%%%%%%%%%%%%%%%%%%%%%%%%%%%%%%%%%%%%%%%%%%%%%

%%%%%%%%%%%%%%%%%%%%%%%%%%%%%%%%%%%%%%%%%%%%%%%%%%%%%%%%%
\begin{figure*}[t!]
    \centering
    \includegraphics[width=\textwidth]{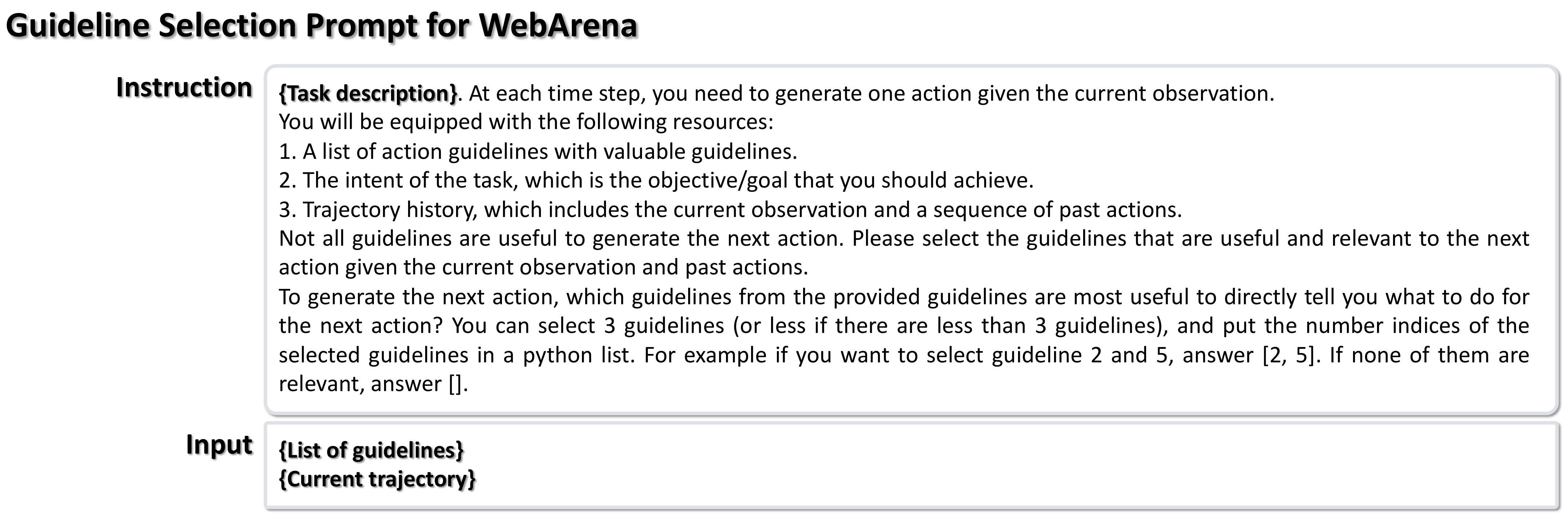}
    \caption{Our prompt template for selecting the most relevant context-aware guidelines during the test time (\Cref{eqn:guideline-selection-module}) in the WebArena domain.}
    \label{fig:guideline-prompt-guide-retrieval-webarena}
\end{figure*}
%%%%%%%%%%%%%%%%%%%%%%%%%%%%%%%%%%%%%%%%%%%%%%%%%%%%%%%%%

For selecting only $k$ most relevant guidelines in case there are more corresponding context-aware guidelines during testing (\Cref{eqn:guideline-selection-module}), we use the prompt with \Cref{fig:guideline-prompt-guide-retrieval-webshop-alfworld} for WebShop and ALFWorld and \Cref{fig:guideline-prompt-guide-retrieval-webarena} for WebArena.

% \clearpage
\section{Example Context-Aware Guidelines}
\begin{figure*}[t!]
    \centering
    \includegraphics[width=\textwidth]{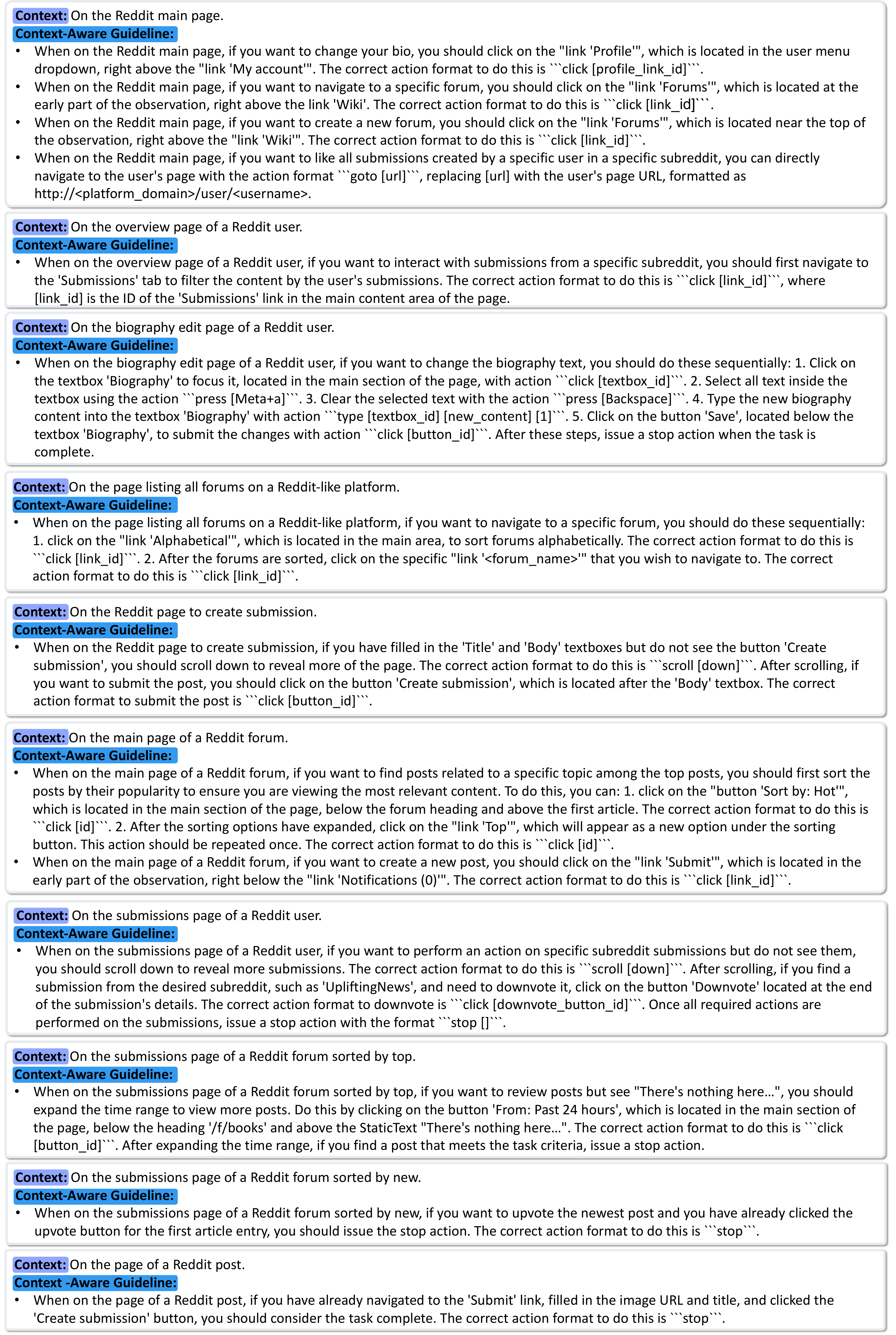}
    \caption{Example contexts and corresponding guidelines for WebArena.}
    \label{fig:state-and-guideline-webarena}
\end{figure*}

\begin{figure*}[t!]
    \centering
    \includegraphics[width=\textwidth]{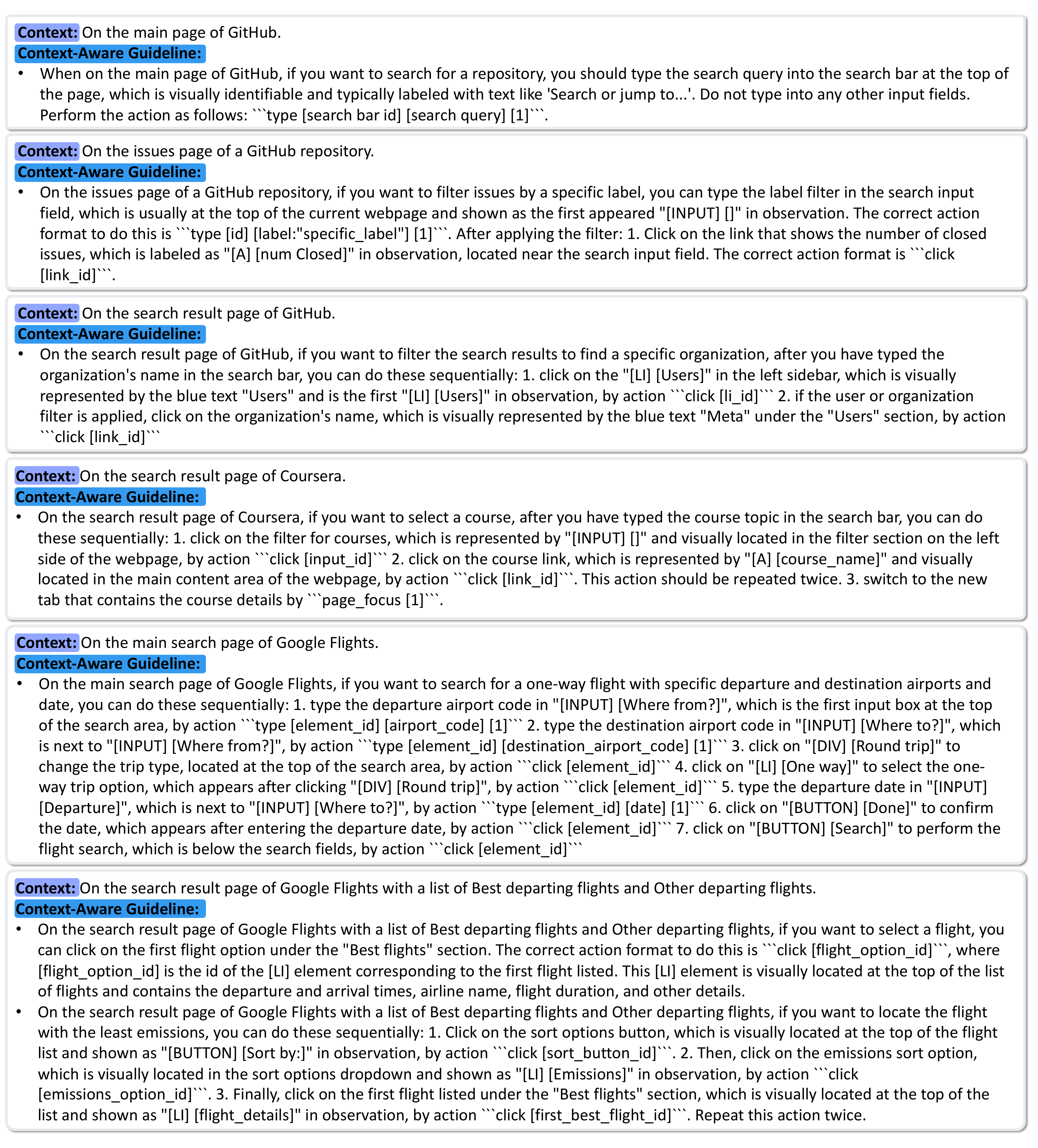}
    \caption{Example contexts and corresponding guidelines for real-world websites.}
    \label{fig:state-and-guideline-real}
\end{figure*}

In \Cref{fig:state-and-guideline-webarena,fig:state-and-guideline-real}, we show a list of possible contexts and context-aware guidelines on WebArena and real-world websites, respectively. 

%%%%%%%%%%%%%%%%%%%%%%%%%%%%%%%%%%%%%%%%%%%%%%%%%%%%%%%%%
%%%%%%%%%%%%%%%%%%%%%%%%%%%%%%%%%%%%%%%%%%%%%%%%%%%%%%%%%

%%%%%%%%%%%%%%%%%%%%%%%%%%%%%%%%%%%%%%%%%%%%%%%%%%%%%%%%%
\newpage
\clearpage
\section*{NeurIPS Paper Checklist}

\begin{enumerate}

\item {\bf Claims}
    \item[] Question: Do the main claims made in the abstract and introduction accurately reflect the paper's contributions and scope?
    \item[] Answer: \answerYes{} % Replace by \answerYes{}, \answerNo{}, or \answerNA{}.
    \item[] Justification: In the abstract and introduction, we clearly outline our main contribution: extracting context-aware guidelines from offline data to enhance the decision-making of LLM-based agents. We support the effectiveness of our method through empirical evaluations in \Cref{sec:evaluation}.

\item {\bf Limitations}
    \item[] Question: Does the paper discuss the limitations of the work performed by the authors?
    \item[] Answer: \answerYes{} % Replace by \answerYes{}, \answerNo{}, or \answerNA{}.
    \item[] Justification: We outline potential limitations of \textsc{\Ours} in \Cref{sec:limitation-broader-impact}.

\item {\bf Theory Assumptions and Proofs}
    \item[] Question: For each theoretical result, does the paper provide the full set of assumptions and a complete (and correct) proof?
    \item[] Answer: \answerNA{} % Replace by \answerYes{}, \answerNo{}, or \answerNA{}.
    \item[] Justification: This paper does not include theoretical contributions.

    \item {\bf Experimental Result Reproducibility}
    \item[] Question: Does the paper fully disclose all the information needed to reproduce the main experimental results of the paper to the extent that it affects the main claims and/or conclusions of the paper (regardless of whether the code and data are provided or not)?
    \item[] Answer: \answerYes{} % Replace by \answerYes{}, \answerNo{}, or \answerNA{}.
    \item[] Justification: \Cref{sec:evaluation-setup} and \Cref{sec:appendix-evaluation-details} provide detailed information (e.g., model versions, hyperparameters, number of training data) for reproducing our experimental results. Additionally, \Cref{sec:appendix-prompt-template} contains our prompt templates.

\item {\bf Open access to data and code}
    \item[] Question: Does the paper provide open access to the data and code, with sufficient instructions to faithfully reproduce the main experimental results, as described in supplemental material?
    \item[] Answer: \answerNo{} % Replace by \answerYes{}, \answerNo{}, or \answerNA{}.
    \item[] Justification: We intend to release the source code upon acceptance.

\item {\bf Experimental Setting/Details}
    \item[] Question: Does the paper specify all the training and test details (e.g., data splits, hyperparameters, how they were chosen, type of optimizer, etc.) necessary to understand the results?
    \item[] Answer: \answerYes{} % Replace by \answerYes{}, \answerNo{}, or \answerNA{}.
    \item[] Justification: We detail our experimental settings in \Cref{sec:evaluation-setup} and \Cref{sec:appendix-evaluation-details}.

\item {\bf Experiment Statistical Significance}
    \item[] Question: Does the paper report error bars suitably and correctly defined or other appropriate information about the statistical significance of the experiments?
    \item[] Answer: \answerNo{} % Replace by \answerYes{}, \answerNo{}, or \answerNA{}.
    \item[] Justification: We provide the average rewards and success rates calculated across test tasks in our results, but we do not include the variance.

\item {\bf Experiments Compute Resources}
    \item[] Question: For each experiment, does the paper provide sufficient information on the computer resources (type of compute workers, memory, time of execution) needed to reproduce the experiments?
    \item[] Answer: \answerYes{} % Replace by \answerYes{}, \answerNo{}, or \answerNA{}.
    \item[] Justification: Our experiments are based on GPT, and we detail the versions of GPT in \Cref{sec:appendix-evaluation-details}.
    
\item {\bf Code Of Ethics}
    \item[] Question: Does the research conducted in the paper conform, in every respect, with the NeurIPS Code of Ethics \url{https://neurips.cc/public/EthicsGuidelines}?
    \item[] Answer: \answerYes{} % Replace by \answerYes{}, \answerNo{}, or \answerNA{}.
    \item[] Justification: Our work conforms the NeurIPS Code of Ethics.

\item {\bf Broader Impacts}
    \item[] Question: Does the paper discuss both potential positive societal impacts and negative societal impacts of the work performed?
    \item[] Answer: \answerYes{} % Replace by \answerYes{}, \answerNo{}, or \answerNA{}.
    \item[] Justification: We provide the broader impacts of our work in \Cref{sec:limitation-broader-impact}.
    
\item {\bf Safeguards}
    \item[] Question: Does the paper describe safeguards that have been put in place for responsible release of data or models that have a high risk for misuse (e.g., pretrained language models, image generators, or scraped datasets)?
    \item[] Answer: \answerNA{} % Replace by \answerYes{}, \answerNo{}, or \answerNA{}.
    \item[] Justification: This paper makes the algorithmic contribution and does not release new data or models.

\item {\bf Licenses for existing assets}
    \item[] Question: Are the creators or original owners of assets (e.g., code, data, models), used in the paper, properly credited and are the license and terms of use explicitly mentioned and properly respected?
    \item[] Answer: \answerYes{} % Replace by \answerYes{}, \answerNo{}, or \answerNA{}.
    \item[] Justification: We reference the original papers and sources throughout the paper.

\item {\bf New Assets}
    \item[] Question: Are new assets introduced in the paper well documented and is the documentation provided alongside the assets?
    \item[] Answer: \answerNA{} % Replace by \answerYes{}, \answerNo{}, or \answerNA{}.
    \item[] Justification: This paper does not release new assets.

\item {\bf Crowdsourcing and Research with Human Subjects}
    \item[] Question: For crowdsourcing experiments and research with human subjects, does the paper include the full text of instructions given to participants and screenshots, if applicable, as well as details about compensation (if any)? 
    \item[] Answer: \answerNA{} % Replace by \answerYes{}, \answerNo{}, or \answerNA{}.
    \item[] Justification: This paper does not involve crowdsourcing nor research with human subjects.

\item {\bf Institutional Review Board (IRB) Approvals or Equivalent for Research with Human Subjects}
    \item[] Question: Does the paper describe potential risks incurred by study participants, whether such risks were disclosed to the subjects, and whether Institutional Review Board (IRB) approvals (or an equivalent approval/review based on the requirements of your country or institution) were obtained?
    \item[] Answer: \answerNA{} % Replace by \answerYes{}, \answerNo{}, or \answerNA{}.
    \item[] Justification: This paper does not involve crowdsourcing nor research with human subjects.

\end{enumerate}

\end{document}